\documentclass{article}

    \PassOptionsToPackage{numbers, compress}{natbib}

\usepackage[main, final]{neurips_2026}

\usepackage[utf8]{inputenc} 
\usepackage[T1]{fontenc}    
\usepackage{hyperref}       
\usepackage{url}            
\usepackage{booktabs}       
\usepackage{amsfonts}       
\usepackage{nicefrac}       
\usepackage{microtype}      
\usepackage{xcolor}         
\usepackage{amsmath}
\usepackage{algorithm}
\usepackage{algorithmic}
\usepackage{pgfplots}
\pgfplotsset{compat=1.18}
\usepgfplotslibrary{colorbrewer}
\usepackage{tabularx}
\usepackage{multirow}
\usepackage{graphicx}
\usepackage{caption}
\usepackage{subcaption}
\usepackage{adjustbox}
\usepackage{wrapfig}
\usepackage[table]{xcolor}
\usepackage{capt-of}
\usepackage{array}

\usepackage{xcolor}
\usepackage{xspace}

\definecolor{pinegreen}{rgb}{0.0, 0.47, 0.44}
\definecolor{cornellred}{rgb}{0.7, 0.11, 0.11}
\definecolor{cadmiumgreen}{rgb}{0.0, 0.42, 0.24}
\definecolor{spirodiscoball}{rgb}{0.06, 0.75, 0.99}
\definecolor{navyblue}{rgb}{0,0.4,0.8}
\definecolor{Red7}{rgb}{0.941, 0.243, 0.243}
\definecolor{Green7}{RGB}{55, 178, 77}
\definecolor{Blue9}{rgb}{0.098,0.3,0.9}
\definecolor{darkred}{rgb}{0.55, 0.0, 0.0}
\definecolor{mygreen}{rgb}{0, 0.6823, 0.7215}
\definecolor{algcommentgray}{gray}{0.45}
\usepackage[table,dvipsnames]{xcolor}

\hypersetup{
  linkcolor = cornellred,
  citecolor  = cadmiumgreen,
  colorlinks = true,
  urlcolor = Blue9
}

\usepackage{pifont}

\newcommand{\yr}[1]{#1}
\newcommand{\methodname}{MVIP\xspace}

\usepackage{amsmath,amsfonts,bm}

\def\1{\bm{1}}

\DeclareMathAlphabet{\mathsfit}{\encodingdefault}{\sfdefault}{m}{sl}
\SetMathAlphabet{\mathsfit}{bold}{\encodingdefault}{\sfdefault}{bx}{n}

\newcommand{\eg}{\textit{e}.\textit{g}.}

\def\onedot{.}
\def\viz{\emph{viz}\onedot}
\def\eg{\emph{e}\onedot\emph{g}\onedot}

\renewcommand{\paragraph}[1]{\vspace{1mm}\noindent\textbf{#1.}\,\,}

\DeclareUnicodeCharacter{27A4}{>}
\DeclareUnicodeCharacter{FFFD}{\$}
\DeclareUnicodeCharacter{25CF}{$\cdot$}
\DeclareUnicodeCharacter{25B6}{>}

\title{Meta Inverse Prompting for Video Generative Models}

\author{%
  Yerin Jung \qquad
  Jongheon Jeong\thanks{Corresponding author} \\
  Korea University \\
  \texttt{\{yerinj,jonghj\}@korea.ac.kr}
}

\begin{document}

\maketitle

\begin{abstract}
{Recent advances in video generative models have enabled flexible conditional generation and editing pipelines, such as text-to-video, image-to-video, and reference-guided synthesis. However, these workflows are inherently one-directional, mapping conditions to videos without providing a mechanism to invert this process. In complex media production settings, the ability to recover controllable textual conditions from a given video is highly desirable, yet remains underexplored. While inverse prompting has been studied for image generation, directly extending these approaches to video is challenging due to the need to capture temporal dynamics beyond static appearance, as well as the substantial computational cost of per-video optimization. In this work, we propose the first practical framework for efficient video inverse prompting. Specifically, we formulate inverse prompting as a meta-learning problem and introduce \emph{Meta Video Inverse Prompting} (\methodname{}), which learns a meta-prompt optimized over temporal dimensions to enable fast and effective prompt inference. Our approach amortizes the inversion process, eliminating the need for costly iterative optimization at test time. Experimental results demonstrate that \methodname{} recovers higher-quality inverse prompts with significantly improved reconstruction fidelity compared to image-based inversion and iterative captioning baselines, while achieving substantial gains in efficiency. Code and video results are available at \url{https://ueriniuno.github.io/mvip-page/}.}

\end{abstract}

\section{Introduction}
{
Video generative models have rapidly advanced to produce high-quality and temporally coherent videos, approaching realistic visual fidelity across diverse scenes and motions~\cite{wang2023modelscope, wan2025wan, yang2025cogvideox}. These models are increasingly deployed across a wide range of applications, including content creation \cite{polyak2024movie, teng2025magi}, editing \cite{yan2023magicprop, zhang2024effived, feng2024ccedit, liu2025anchorsync}, simulation \cite{bruce2024genie, yang2023unisim}, and world modeling \cite{alonso2024diffusion,bruce2024genie, chen2025learning}. To support such applications, modern video generation frameworks support flexible conditional pipelines such as text-to-video \cite{ singer2023makeavideo, wang2023modelscope, yang2025cogvideox}, image-to-video \cite{blattmann2023stable, chen2024videocrafter2, wan2025wan}, and reference-guided synthesis \cite{li2023videogen, zhou2026scaling}.}


{
However, despite this progress, existing workflows to access video generative models are inherently ``unidirectional''; \viz, while conditions can effectively guide video synthesis, they lack a clear mechanism to invert this process, \eg, for fine-grained semantic controls. For example, a creator may generate a video from a prompt and later wish to modify it, such as changing lighting or motion while preserving the scene. However, the original prompt is often unavailable or insufficient to reproduce the video due to the non-unique and model-dependent nature of prompt-video mappings, forcing users to rely on trial-and-error prompt engineering. 
}

{
This motivates the problem of \emph{inverse prompting}, which aims to recover a textual prompt that can faithfully reproduce a given target video under a generative model. Prior work on inverse prompting has largely focused on images \cite{wen2023hard, mahajan2024prompting, ren2025reverse}, primarily relying on gradient-based optimization or iterative captioning strategies. However, these approaches do not scale well to video. Gradient-based methods \cite{wen2023hard, mahajan2024prompting} require repeated forward and backward passes during target-specific prompt optimization, which becomes prohibitively expensive for long video sequences. Captioning-based approaches \citep{ren2025reverse}, on the other hand, often produce semantically plausible but insufficient prompts to support reconstruction, \yr{as videos require capturing not only per-frame appearance but also temporal dynamics and coherence across frames,} \eg, action sequences, motion trajectories, and camera movements. 
Consequently, directly applying existing methods to video results in a poor trade-off between reconstruction fidelity and computational efficiency, while the problem itself remains largely unexplored in the video domain. As shown in Figure~\ref{fig:overview}, our goal is to recover a reusable prompt from a target video, enabling both faithful video reconstruction and controllable prompt-based video editing.


\begin{figure}[t]
    \centering
    \includegraphics[width=\linewidth]{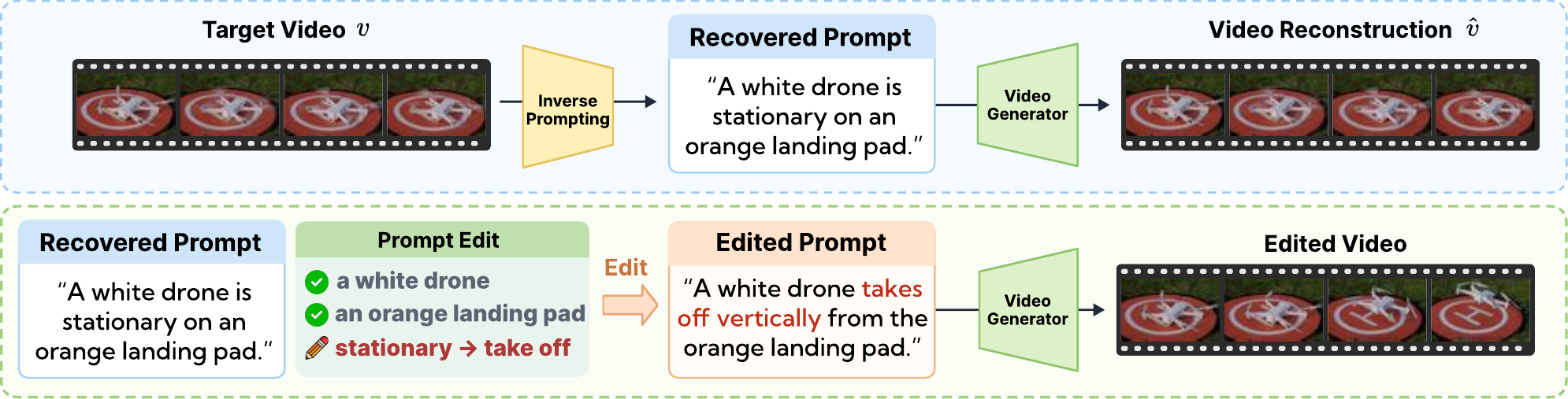}
    \caption{
    \yr{\textbf{Illustration of \methodname{} and downstream editing.}}
{\methodname{} recovers a prompt from a target video through inverse prompting, enabling a video generator conditioned on the prompt to reproduce the video. Using the recovered prompt, video generation can be controlled by editing the prompt, enabling controllable video editing.}}
\vspace{-0.1in}

    \label{fig:overview} 
\end{figure}

\paragraph{Contributions}
{In this paper, we propose \emph{Meta Video Inverse Prompting} (\methodname{}), the first practical framework for video inverse prompting, to the best of our knowledge. Our key idea is to unify gradient-based optimization and captioning-based approaches within a meta-learning~\cite{finn2017model, nichol2018first} formulation, achieving both high reconstruction fidelity and efficiency. Instead of optimizing prompts from scratch for each video, we cast inverse prompting as a meta-learning that learns a shared prompt initialization across videos. During meta-training, \methodname{} performs gradient-based prompt optimization over diverse videos, allowing the meta-prompt to capture common spatiotemporal structures required for faithful reconstruction. At inference time, this learned initialization enables rapid prompt inference without costly per-video optimization, effectively behaving as a reconstruction-aware captioning mechanism. In this way, \methodname{} combines the reconstruction fidelity of gradient-based methods with the efficiency of captioning-based approaches. 

We evaluate \methodname{} on both real and synthetic videos, including videos generated from diverse video generative models, to assess its generality across different video distributions and generation pipelines. Since video inverse prompting remains largely unexplored and lacks established benchmarks or baselines, we construct practical comparison methods based on recent image inverse prompting approaches, including PEZ~\cite{wen2023hard} and ARPO~\cite{ren2025reverse}, as well as captioning-based methods. The results demonstrate that \methodname{} consistently achieves superior reconstruction performance across diverse video-video and video-text alignment metrics. Notably, our method significantly outperforms iterative captioning approaches despite requiring substantially lower computational cost, showing that high-quality inverse prompting cannot be achieved by simply increasing inference-time iterations. \yr{Qualitative comparisons further show that our meta-learned initialization produces prompts that better capture temporal dynamics and fine-grained scene details, resulting in more faithful and controllable video reconstructions.}

\vspace{-0.05in}
\section{Related Work}
\vspace{-0.05in}
\paragraph{Video generative models}
Recent advances in video generation have been largely driven by text-to-video (T2V) frameworks, where diffusion models~\cite{blattmann2023align,wang2023modelscope, wan2025wan}, transformer architectures~\cite{hong2023cogvideo,
gupta2024photorealistic}, and flow-matching objectives~\cite{
kong2024hunyuanvideo,polyak2024movie} have enabled high-quality video synthesis from textual descriptions. Alongside open research models~\cite{chen2024videocrafter2,
kong2024hunyuanvideo,lin2024open,wan2025wan,yang2025cogvideox}, commercial systems~\cite{openai2024sora,runway2024gen3,google2025veo} have further established video generation as a practical tool for content creation. This progress has extended video generation beyond T2V to conditional interfaces such as image-to-video generation \cite{zhang2023i2vgen,liu2024physgen,hu2024animate,ma2025consistent}, reference-guided synthesis \cite{jiang2024videobooth,xu2026smrabooth}, and text-driven editing~\cite{qi2023fatezero,liu2024video,wang2025taming}. Here, prompts serve not only as semantic descriptions but also as control signals that guide visual content, motion, and editing.

\paragraph{Prompt optimization and inversion}
Recent video generative models have achieved substantial progress, yet their outputs remain highly sensitive to input conditions, including textual prompts and visual anchors. Minor variations in these conditions can lead to substantially different visual outputs \cite{ji2025prompt,zhang2025rise}. Existing prompt inversion methods have mainly focused on images, including pseudo-word embedding learning \cite{gal2023an}, model fine-tuning with unique identifiers \cite{ruiz2023dreambooth},
gradient-based hard-prompt optimization \cite{wen2023hard,mahajan2024prompting}, and black-box prompt search \cite{ren2025reverse}. While effective for static visual content, these methods are not directly suited to video, where the recovered prompt must account for temporal evolution in addition to appearance. In this paper, we address this inverse problem through a reconstruction-driven framework that recovers prompts directly from target videos.

\paragraph{Meta-learning}
Meta-learning aims to extract transferable knowledge from a distribution of tasks so that a model can rapidly adapt to new tasks with limited data or optimization. Beyond metric-based methods~\cite{vinyals2016matching} that learn shared embeddings and model-based approaches~\cite{santoro2016meta} that leverage memory, optimization-based meta-learning is a prominent paradigm. Notably, MAML~\cite{finn2017model} and Reptile~\cite{nichol2018first} focus on learning an optimal initialization for rapid gradient-based adaptation. This paradigm has been widely adopted in domains where task-specific adaptation is costly, such as personalized federated learning~\cite{fallah2020personalized}, few-shot image classification~\cite{li2017meta, rusu2018metalearning}, and natural language classification~\cite{bansal2020learning}. A similar initialization-based view has also been explored in prompt learning and prompt optimization. Meta prompt tuning learns transferable initializations for cross-task generalization~\cite{qin2023learning}, while system prompt optimization uses a meta-learning framework to learn system prompts that generalize across diverse user inputs and unseen tasks~\cite{choi2025system}. These studies highlight the benefit of transferring task-level experience through learned prompt initializations. Building on this, we formulate video inverse prompting as a meta-learning problem, leveraging shared initializations to facilitate precise, reconstruction-oriented adaptation for each target video.

\vspace{-0.05in}
\section{Method}
\vspace{-0.05in}

{In this section, we introduce \emph{Meta Video Inverse Prompting} (\methodname{}), the first practical framework for video inverse prompting, to the best of our knowledge. Our key insight is that gradient-based prompt optimization for video reconstruction can be lifted into the instruction-prompt space within a vision-language model (VLM) and amortized via meta-learning, enabling it to directly generate reconstruction-aware prompts from unseen videos in a fast, captioning-like manner.}

{We first formalize our problem of interest, \viz, video inverse prompting, in Section~\ref{sec:setup}. In Section~\ref{sec:background}, we provide a brief overview of the meta-learning framework underlying our approach. Finally, we present the details of our proposed method, \methodname{}, in Section~\ref{sec:meta}.
An overview of the proposed framework is illustrated in Figure~\ref{fig:meta}.}

\subsection{Problem setup}
\label{sec:setup}
\begin{figure}[t]
    \centering
    \includegraphics[width=\linewidth]{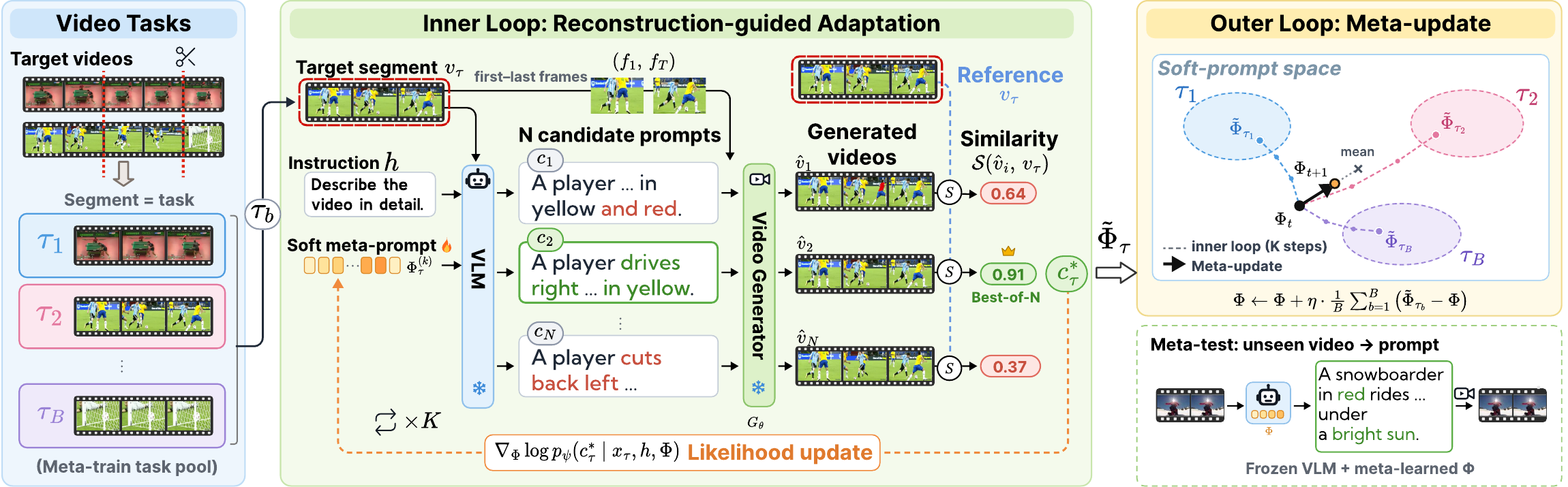}
    \caption{\textbf{Overview of \methodname{}.} 
    \methodname{} meta-learns a shared soft-prompt initialization for video inverse prompting across video tasks. During inner-loop adaptation, a frozen VLM generates candidate prompts, and the candidate with the highest reconstruction score from a frozen video generator is selected as a pseudo-target for updating the soft prompt. The task-adapted prompts are then used to update the shared initialization, which can be directly applied to unseen videos.
    }
    \label{fig:meta}
\end{figure}
We consider a conditional video generative model \(G\), conditioned on a textual prompt \(c\) and a first-last-frame condition 
\(v_{\mathrm{cond}} = (f_1^v, f_T^v)\), where \(f_1^v\) and \(f_T^v\) denote the first and last frames of \(v\), respectively. Our goal is to recover a textual prompt $c$ that enables a video generator \(G\) to reconstruct the target video $v$. Since text alone underconstrains video reconstruction, we use the first and last frames to specify how the video starts and ends. We write the generator output for a candidate prompt as 
\(\hat{v} = G_{\theta}(c, v_{\mathrm{cond}})\). Video inverse prompting then seeks the prompt whose generated video best reconstructs the target video:
\begin{equation}
c^{*}(v) = \arg\max_{c}
S\bigl(G(c, v_{\mathrm{cond}}), v\bigr).
\label{eqn:meta_learning_objective}
\end{equation}
Here, \(S(\cdot,\cdot)\) denotes a video-level reconstruction similarity score between the generated and target videos. In our setting, \(G\) is a fixed black-box generator, so we do not update its parameters or access its internal gradients. Instead, we solve video inverse prompting by searching for a textual prompt \(c\) that maximizes the reconstruction score under the fixed first-last-frame condition \(v_{\mathrm{cond}}\).

\subsection{Background: Meta-learning} 
\label{sec:background} 
\yr{Meta-learning aims to acquire transferable knowledge that enables a learner to quickly adapt to new tasks. It consists of meta-training and meta-testing. During meta-training,
the learner uses multiple tasks to learn transferable knowledge across tasks. At meta-test time, the learned knowledge is used to facilitate adaptation to an unseen task.}



In gradient-based meta-learning, this transferable knowledge is commonly represented as an initialization $\theta$. Let $A_{\tau}^{K}(\theta)$ denote the parameters obtained after $K$ task-specific adaptation steps from $\theta$ on
a task $\tau$ sampled from the task distribution $p(\tau)$. 
A representative approach in this direction is \emph{Model-Agnostic Meta-Learning} (MAML) \cite{finn2017model}, which optimizes the initialization by differentiating the post-adaptation validation loss through the adaptation process:
$ 
\theta
\leftarrow
\theta
-
\eta \cdot 
\mathbb{E}_{\tau}\left[
\nabla_{\theta}
\mathcal{L}_{\tau}
\left(
A_{\tau}^{K}(\theta)
\mid
\mathcal{D}_{\tau}^{\mathrm{val}}
\right) \right]
$, 
where 
$\eta$ is the meta-learning rate, and $\mathcal{D}_{\tau}^{\mathrm{val}}$ denotes the validation data used to evaluate the task-adapted parameters.

Reptile~\cite{nichol2018first} follows a similar task-adaptation procedure but adopts a significantly simpler meta-update rule. In contrast to MAML~\citep{finn2017model}, which computes the meta-gradient by differentiating through task-specific adaptation and thus involves second-order derivatives, Reptile~\cite{nichol2018first} avoids differentiating through the adaptation trajectory. Specifically, it directly updates the initialization toward the task-adapted parameters using a first-order meta-update:
\begin{equation}
\theta
\leftarrow
\theta
+
\eta \cdot 
\mathbb{E}_\tau
\left[
A_{\tau}^{K}(\theta) - \theta
\right].
\label{eq:reptile_update}
\end{equation}
In our setting, differentiating through the adaptation process is particularly challenging, since video generation is computationally expensive and prompt generation involves discrete captioning processes within a VLM. Therefore, our approach is instead motivated by Reptile~\cite{nichol2018first},
which performs meta-learning through simple first-order updates toward task-adapted parameters. Building on this intuition, we next introduce our meta-learning formulation for video inverse prompting.


\subsection{\methodname{}: Meta Video Inverse Prompting}
\label{sec:meta}

We adopt a meta-learning perspective for video inverse prompting by treating each target video segment as a task and learning a prompt-space initialization. In \methodname{}, the meta-learned initialization $\theta$ introduced in
Section~\ref{sec:background} is instantiated as a VLM-side soft prompt $\Phi$. For each task $\tau$, $\Phi$ is adapted for $K$ steps to the target video, and the resulting task-adapted soft prompt is used to update the shared initialization following Reptile~\citep{nichol2018first}. The resulting meta-learned soft prompt $\Phi$ can be directly applied to unseen video segments. \methodname{} optimizes only $\Phi$, while keeping both the VLM $M_\psi$ and video generator $G$ frozen.


\paragraph{Prompt-space parameterization}
The prompt $c$ in Eq.~\ref{eqn:meta_learning_objective} denotes the discrete textual prompt used by the video generator $G$, whereas the learnable soft prompt $\Phi$ conditions the VLM $M_\psi$. For task $\tau$, let
$x_\tau=\{f_r^{v_\tau}\}_{r\in\mathcal R}$ denote the frames sampled from $v_\tau$ as input to the VLM, and let
$v_{\tau,\mathrm{cond}}=(f_1^{v_\tau},f_T^{v_\tau})$ denote its first-last-frame condition. Let $h$ denote the fixed hard instruction used for prompt generation. We write
\begin{equation}
c_\tau(\Phi)
\sim
p_{\psi}
\!\left(
\cdot \mid x_\tau,h,\Phi
\right),
\qquad
\hat{v}_\tau
=
G
\!\left(
c_\tau(\Phi), v_{\tau,\mathrm{cond}}
\right).
\label{eq:prompt_space_parameterization}
\end{equation}
Thus, \methodname{} steers prompt generation through the soft prompt $\Phi$, while the resulting hard prompt $c_\tau(\Phi)$ remains in the textual space compatible with the video generator.

\paragraph{Reconstruction-guided inner adaptation}
For each task $\tau$, we initialize the task-specific soft prompt as $\Phi_{\tau}^{(0)}=\Phi$. At inner step $k$, we sample $N$ candidate prompts
$c_{\tau,i}^{(k)} \sim
p_\psi(\cdot \mid x_\tau,h,\Phi_\tau^{(k-1)})$
and select the candidate with the highest reconstruction score:
\begin{equation}
i^*
=
\arg\max_{i=1,\ldots,N}
S
\!\left(
G(c_{\tau,i}^{(k)}, v_{\tau,\mathrm{cond}}), v_\tau
\right),
\qquad
c_{\tau}^{*,(k)}
=
c_{\tau,i^*}^{(k)} .
\label{eq:bestofn_selection}
\end{equation}
The selected prompt $c_{\tau}^{*,(k)}$ serves as a reconstruction-guided pseudo-target for updating the soft prompt. Since the reconstruction score is used only for candidate selection,
we adapt the soft prompt by maximizing the VLM likelihood of the selected prompt:
\begin{equation}
\Phi_{\tau}^{(k)}
=
\Phi_{\tau}^{(k-1)}
+
\alpha
\nabla_{\Phi_{\tau}^{(k-1)}}
\log
p_{\psi}
\!\left(
c_{\tau}^{*,(k)}
\mid x_\tau,h,\Phi_{\tau}^{(k-1)}
\right).
\label{eq:inner_likelihood_update}
\end{equation}
After $K$ steps, we obtain the task-adapted soft prompt
$\tilde{\Phi}_{\tau}=\Phi_{\tau}^{(K)}=A_{\tau}^{K}(\Phi)$.

\paragraph{Meta-update}
The outer loop aggregates the task-adapted soft prompts into a shared prompt-space initialization. At iteration $t$, inner adaptation starts from $\Phi_t$ for each task in a batch $\{\tau_b\}_{b=1}^{B}$, producing $\tilde{\Phi}_{t,b}=A_{\tau_b}^{K}(\Phi_t)$. We then update the shared initialization toward the average task-adapted soft prompt:
\begin{equation}
\Phi_{t+1}
=
\Phi_t
+
\eta
\left[
\frac{1}{B}
\sum_{b=1}^{B}
\left(
\tilde{\Phi}_{t,b}-\Phi_t
\right)
\right].
\label{eq:meta_update}
\end{equation}
This update amortizes reconstruction-guided adaptation across tasks into a single prompt-space initialization. At inference time, the learned initialization can be directly used to
condition the VLM on unseen video segments, producing reconstruction-oriented prompts for the video generator. The complete meta-training procedure is summarized in Algorithm~\ref{alg:bestofn_reptile} in Appendix.

\section{Experiments}
\begin{table}[t]
\centering
\caption{
\textbf{Quantitative comparison of MVIP against baselines.}
\yr{Performance is evaluated across video-level, visual, and image-alignment metrics on both synthetic and real-world videos. \textbf{Bold} and \underline{underline} denote the best and runner-up results, respectively. \methodname{} achieves the best overall reconstruction in both domains.
}}
\label{4}
\label{tab:main_table}
\small
\setlength{\tabcolsep}{4.5pt}
\renewcommand{\arraystretch}{1.08}

\begin{adjustbox}{width=\textwidth}
\begin{tabular}{lcccccc}
\toprule
& \multicolumn{3}{c}{Video Similarity}
& \multicolumn{2}{c}{Visual Similarity}
& \multicolumn{1}{c}{Alignment} \\
\cmidrule(lr){2-4}
\cmidrule(lr){5-6}
\cmidrule(lr){7-7}

Method
& ViCLIP V-V$\uparrow$
& VideoMAE$\uparrow$
& RAFT Dist.$\downarrow$
& DINOv2$\uparrow$
& DreamSim$\downarrow$
& CLIP-I$\uparrow$ \\

\midrule
\multicolumn{7}{c}{\textbf{Meta-training datasets}} \\
\midrule

\multicolumn{7}{l}{
\textit{Synthetic videos}
{\scriptsize
(from VC2 \cite{chen2024videocrafter2},
MS \cite{wang2023modelscope},
Pika \cite{pika2023},
and Wan 2.2 \cite{wan2025wan})
}
} \\

FLF-Only
& \underline{0.9424}
& 0.9878
& \underline{0.3777}
& \underline{0.8678}
& \underline{0.0667}
& \underline{0.9272} \\

PEZ~\cite{wen2023hard}
& 0.9381
& 0.9860
& 0.3820
& 0.8583
& 0.0747
& 0.9255 \\

Captioning (Qwen2.5-VL~\cite{bai2025qwen25vl})
& 0.9284
& 0.9828
& 0.3802
& 0.8505
& 0.0897
& 0.9233 \\

Captioning + Refine~\cite{ren2025reverse}
& 0.9327
& \underline{0.9880}
& 0.4288
& 0.8576
& 0.0775
& 0.9261 \\

\rowcolor{blue!8}
\textbf{MVIP} (\textbf{Ours})
& \textbf{0.9460}
& \textbf{0.9905}
& \textbf{0.3530}
& \textbf{0.8785}
& \textbf{0.0636}
& \textbf{0.9335} \\

\midrule

\multicolumn{7}{l}{
\textit{Real videos}
{\scriptsize
(from ActivityNet \cite{caba2015activitynet}
and MSR-VTT \cite{xu2016msr})
}
} \\

FLF-Only
& 0.6992
& 0.9474
& \underline{0.5772}
& 0.5116
& 0.3421
& 0.7324 \\

PEZ~\cite{wen2023hard}
& 0.7883
& \textbf{0.9536}
& 0.5857
& \underline{0.5710}
& \underline{0.3166}
& 0.7643 \\

Captioning (Qwen2.5-VL~\cite{bai2025qwen25vl})
& 0.7977
& 0.9488
& 0.5955
& 0.5705
& 0.3300
& 0.7668 \\

Captioning + Refine~\cite{ren2025reverse}
& \underline{0.8047}
& 0.9489
& 0.6040
& 0.5618
& 0.3295
& \underline{0.7712} \\

\rowcolor{blue!8}
\textbf{MVIP} (\textbf{Ours})
& \textbf{0.8328}
& \underline{0.9523}
& \textbf{0.5738}
& \textbf{0.5848}
& \textbf{0.3121}
& \textbf{0.7825} \\

\midrule
\multicolumn{7}{c}{\textbf{Unseen datasets}} \\
\midrule

\multicolumn{7}{l}{
\textit{Kinetics-400}
{\scriptsize
\cite{kay2017kinetics}
}
} \\

FLF-Only
& 0.8808
& \underline{0.9807}
& \textbf{0.4880}
& 0.7483
& 0.1620
& 0.8598 \\

PEZ~\cite{wen2023hard}
& 0.8963
& 0.9776
& 0.5058
& 0.7659
& 0.1619
& 0.8678 \\

Captioning (Qwen2.5-VL~\cite{bai2025qwen25vl})
& \underline{0.9007}
& 0.9803
& 0.4932
& \underline{0.7753}
& \underline{0.1533}
& \underline{0.8712} \\


\rowcolor{blue!8}
\textbf{MVIP} (\textbf{Ours})
& \textbf{0.9186}
& \textbf{0.9829}
& \underline{0.4897}
& \textbf{0.7933}
& \textbf{0.1419}
& \textbf{0.8804} \\

\midrule

\multicolumn{7}{l}{
\textit{DAVIS-2017}
{\scriptsize
\cite{perazzi2016benchmark,pont20172017}
}
} \\

FLF-Only
& 0.9364
& 0.9617
& 0.4320
& 0.8018
& 0.1374
& 0.9074 \\

PEZ~\cite{wen2023hard}
& 0.9451
& \underline{0.9669}
& \underline{0.4097}
& 0.8132
& 0.1411
& 0.9152 \\

Captioning (Qwen2.5-VL~\cite{bai2025qwen25vl})
& \underline{0.9525}
& 0.9651
& 0.4105
& \underline{0.8278}
& \underline{0.1348}
& \underline{0.9210} \\


\rowcolor{blue!8}
\textbf{MVIP} (\textbf{Ours})
& \textbf{0.9568}
& \textbf{0.9790}
& \textbf{0.3412}
& \textbf{0.8364}
& \textbf{0.1282}
& \textbf{0.9234} \\

\bottomrule
\end{tabular}
\end{adjustbox}
\end{table}

In this section, we evaluate \methodname{} on both real and synthetic video datasets. We compare \methodname{} with caption-based and image inverse prompting baselines to assess reconstruction fidelity, and analyze how meta-learned initialization affects the efficiency and quality of the reconstruction prompt. \yr{We also analyze the iterative refinement baseline to examine whether feedback-based updates provide additional gains over one-shot prompt generation.}

\paragraph{Implementation details}
We use a Wan2.2~\cite{wan2025wan}-based first/last-frame conditioned video generator for all reconstruction experiments. Long-form videos are split into scene-consistent segments with PySceneDetect, and first/last-frame conditioning is applied to each segment. This segment-level protocol is shared by all methods to keep generation and evaluation consistent. We use Qwen2.5-VL-7B-Instruct~\cite{bai2025qwen25vl} as the VLM for captioning and comparison, while keeping both the VLM and video generator frozen. Full implementation details are provided in
Appendix~\ref{appen:implementation_details}.

\paragraph{Datasets}
\label{datasets}
We evaluate on both real-world and synthetic videos. The real-world split consists of ActivityNet~\citep{caba2015activitynet} and MSR-VTT~\citep{xu2016msr}, covering long human-activity videos and short everyday web videos, respectively. The synthetic split is built from VidProM~\citep{wang2024vidprom}, a large-scale text-to-video prompt-gallery dataset containing videos generated by Pika~\citep{pika2023},
VideoCrafter2~\citep{chen2024videocrafter2}, and
ModelScope~\citep{wang2023modelscope}, along with
Wan2.2~\citep{wan2025wan} videos generated using VidProM prompts. \yr{We additionally evaluate on Kinetics-400~\citep{kay2017kinetics} and DAVIS-2017~\citep{perazzi2016benchmark,pont20172017} to assess generalization beyond the meta-training datasets.}

\paragraph{Baselines}
\label{baselines}
We compare \methodname{} with four baselines. FLF-Only uses boundary-frame visual anchors without text. PEZ~\cite{wen2023hard} applies image inverse prompting to four sampled frames and concatenates the recovered prompts. Captioning uses the VLM instruction ``Describe this video in detail.'' Captioning+Refine adapts ARPO~\cite{ren2025reverse}, an image reverse prompting method, to video by refining prompts through comparisons between sampled target and generated frames. All baselines follow the same evaluation protocols as \methodname{}, with details in Appendix~\ref{appen:baselines}.

\subsection{Main results}

\begin{figure}[t]
    \centering
    \includegraphics[width=\linewidth]{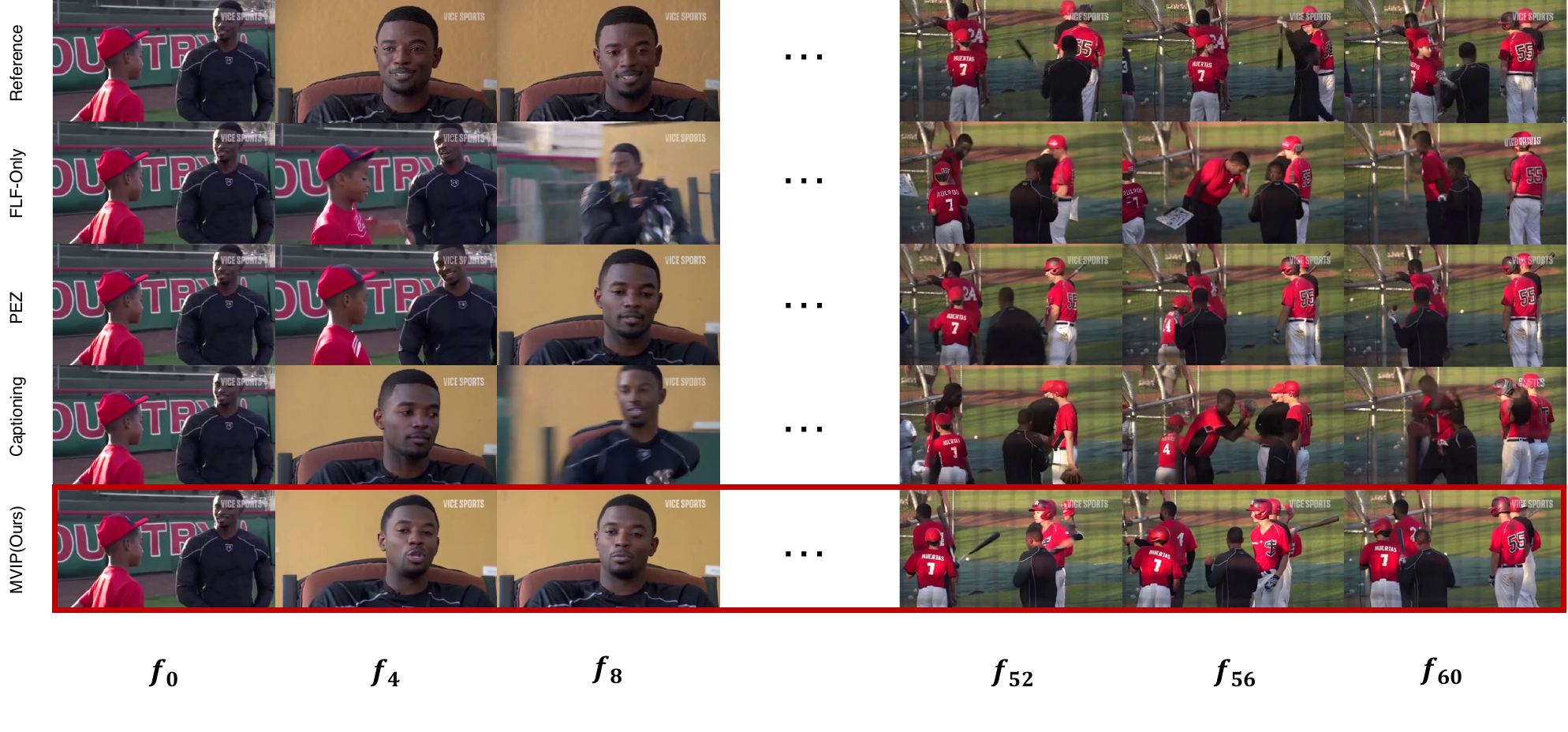}
    \vspace{-8mm}
    \caption{
    \textbf{Qualitative comparison of video reconstruction over time.}
    Compared with the baselines, which begin to drift from the reference video at early time steps such as $f=4$ or $f=8$, \methodname{} preserves the reference content and temporal progression throughout the sequence, remaining consistent even at $f=60$. Additional discussion is provided in Appendix~\ref{appen:recon_qual}.
    }
    \label{fig:app_tab1_2}
\end{figure}

\begin{table}[t]
    \centering
    \caption{\textbf{Image-text qualitative comparison.}
    {Given sampled frames from the target video, our prompt inversion produces more temporally informative and semantically coherent descriptions compared to captioning and image-based inverse prompting baselines.}}
    \scriptsize
    \setlength{\tabcolsep}{3pt}
    \renewcommand{\arraystretch}{1.15}
    \vspace{0.05in}
    \begin{tabularx}{\textwidth}{
        @{}>{\raggedright\arraybackslash}p{0.15\textwidth}
        *{6}{>{\centering\arraybackslash}X}@{}
    }
        \toprule

        \raisebox{0.42in}{\textbf{Target Video}} &
        \includegraphics[width=\linewidth]{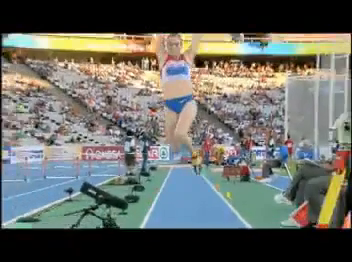} &
        \includegraphics[width=\linewidth]{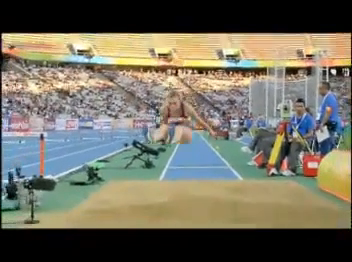} &
        \includegraphics[width=\linewidth]{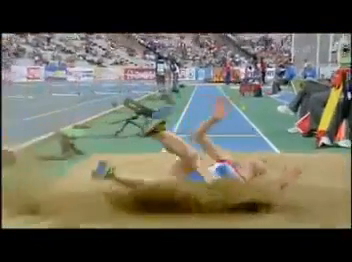} &
        \includegraphics[width=\linewidth]{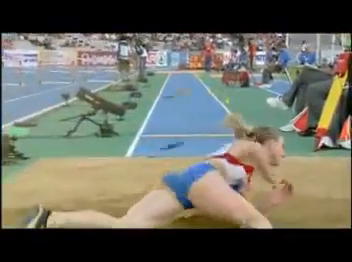} &
        \includegraphics[width=\linewidth]{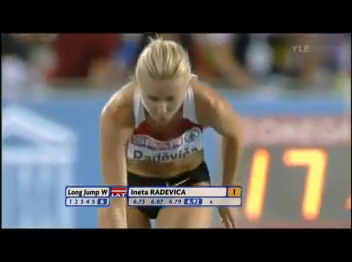} &
        \includegraphics[width=\linewidth]{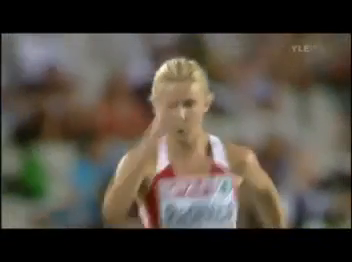} \\
        \cmidrule(lr){2-7}

        \textbf{Sampled Frame} &
        $f_1$ & $f_2$ & $f_3$ & $f_4$ & $f_5$ & $f_6$ \\
        \midrule
        
        Qwen2.5-VL~\cite{bai2025qwen25vl} &
        \multicolumn{6}{>{\raggedright\arraybackslash}p{0.82\textwidth}@{}}{
        A female athlete is running down the track, preparing to jump into a sand pit for long jump competition. She leaps off her feet and lands gracefully on one foot before landing fully onto both feet inside of the sandpit. The camera then cuts away from the action as she walks back towards where she started.
        } \\
        \addlinespace[4pt]

        PEZ~\cite{wen2023hard} &
        \multicolumn{6}{>{\raggedright\arraybackslash}p{0.82\textwidth}@{}}{
        1. ipl lick minogue javelin athletics hurdle musicvideo \ldots{}
        2. cracking gopher crumbs ibm dden euro decathathletics \ldots{}
        3. hewadrenaline sprinter jumper latvia serbia \ldots{}
        4. robyn gwynecapped athletics athletics equalsardinia \ldots{}
        } \\
        \addlinespace[4pt]
        
        \textbf{MVIP (Ours)} &
        \multicolumn{6}{>{\raggedright\arraybackslash}p{0.82\textwidth}@{}}{
        \textbf{
        The video captures a female athlete in the midst of a long jump event at an
        indoor stadium. \ldots{} She is running towards the sand pit, her body leaning
        forward as she approaches the takeoff board. As she reaches the board, she
        leaps into the air. \ldots{} Her body is in mid-air when the video cuts to a
        close-up of her face. She lands in the sand pit. \ldots{} The video then cuts
        to a scoreboard displaying her name, "Ineta Radevica," and her jump distance
        of 6.97 meters.
        }
        } \\

        \bottomrule
    \end{tabularx}
    \label{image-text}
\end{table}

\paragraph{Quantitative results}
\yr{In Table~\ref{tab:main_table}, we first evaluate \methodname{} on held-out videos from datasets used during meta-training, covering both synthetic and real-world videos. \methodname{} achieves the best overall performance in both domains.} Specifically, in the synthetic video subset, \methodname{} consistently achieves higher scores than existing methods, including VLM-based captioning and iterative refinement. In the real-world video domain, \methodname{} increases the ViCLIP V-V score to 0.8328, compared to 0.8047 achieved by the 3-step iterative refinement baseline. \yr{To further evaluate generalizability beyond the meta-training datasets, we evaluate \methodname{} on two unseen test sets, Kinetics-400~\citep{kay2017kinetics} and DAVIS-2017~\citep{perazzi2016benchmark,pont20172017}, without changing the learned meta-prompt or any other training procedure. \methodname{} achieves the best performance on most metrics across both datasets, suggesting that the learned meta-prompt generalizes well to previously unseen video distributions. As shown in Table~\ref{tab:iteration-cost}, iterative refinement provides only modest gains at substantially higher computational cost while remaining below \methodname{} in reconstruction quality. Overall, these results show that \methodname{} improves the quality-efficiency trade-off by reducing reliance on costly test-time refinement. We further report text-alignment results in Appendix~\ref{appen:it_align}.}

\paragraph{Qualitative results}
As shown in Figure~\ref{fig:app_tab1_2}, \methodname{} reconstructs the target video more faithfully than the baselines, preserving the subject, scene composition, and temporal progression across sampled frames. In contrast, FLF-Only and PEZ~\citep{wen2023hard} often drift from the target at corresponding time points, while captioning captures only coarse semantics and produces inconsistent visual content. Overall, the proposed approach better aligns with the target over time while maintaining stronger temporal consistency across frames. Table~\ref{image-text} shows that \methodname{} recovers a more readable and detailed prompt, capturing the action, appearance, scene context, and temporal progression better than other baselines. \yr{We further examine the effect and interpretation of the learned soft prompt in Appendix~\ref{appen:soft_prompt_analysis}.}

\paragraph{Runtime analysis} 
We evaluate whether test-time refinement makes captioning competitive with \methodname{}. We adapt a black-box image inverse-prompting method to video, where the initial VLM caption is iteratively rewritten using feedback from the generated video. \yr{Runtime is reported as the average seconds per video-second across real and synthetic test videos.} As shown in Table~\ref{tab:iteration-cost}, ``Captioning + Refine'' yields modest gains but incurs a high computational cost: compared with \methodname{} at $K=0$, it requires 3.88$\times$ the runtime while still underperforming on both metrics. Per-iteration results are in Appendix~\ref{appen:quan}.

\paragraph{\yr{Test-time adaptation}}
\yr{To assess the effect of test-time adaptation, we evaluate $K \in \{0,1,2\}$ reconstruction-guided adaptation steps per video, starting from the same meta-learned $\Phi$. As shown in Table~\ref{tab:iteration-cost}, even without test-time adaptation ($K=0$), \methodname{} already outperforms all competing baselines on both metrics. A single adaptation step achieves the best reconstruction quality
(0.8392 ViCLIP V-V and 0.7861 CLIP-I), while all evaluated values of $K$ outperform the baselines on both metrics.}

\begin{table*}[t]
\centering

\begin{minipage}[t]{0.54\textwidth}
    \centering
    \captionof{table}{
    \textbf{Runtime and reconstruction quality comparison.}
    \yr{
    Reconstruction quality is evaluated on real-world videos. Inference time is measured on a single NVIDIA B200 GPU instance with two Intel Xeon 6960P CPUs (72 cores each). \textbf{Bold} and \underline{underline} denote the best and runner-up results, respectively.
    }}
    \label{tab:iteration-cost}
    \small
    \setlength{\tabcolsep}{4pt}
    \renewcommand{\arraystretch}{1.1}

    \resizebox{\linewidth}{!}{
    \begin{tabular}{lccc}
        \toprule
        Method
        & Runtime (s/video-second) $\downarrow$
        & ViCLIP V-V $\uparrow$
        & CLIP-I $\uparrow$ \\
        \midrule

        PEZ
        & 342.7 $\pm$ 1.9
        & 0.7883
        & 0.7643 \\

        Captioning (Qwen2.5-VL)
        & \textbf{261.9} $\pm$ 1.2
        & 0.7977
        & 0.7668 \\

        Captioning + Refine
        & 1060.0 $\pm$ 3.7
        & 0.8047
        & 0.7712 \\

        \cmidrule[0.3pt](lr){1-4}
        
        \rowcolor{blue!8}
        \multicolumn{4}{l}{\textbf{\methodname{} (Ours)}} \\

        \hspace{1em}$K=0$ (no adaptation)
        & \underline{273.1} $\pm$ 1.0
        & \underline{0.8328}
        & \underline{0.7825} \\
        
        \hspace{1em}$K=1$
        & 1082.8 $\pm$ 1.7
        & \textbf{0.8392}
        & \textbf{0.7861} \\
        
        \hspace{1em}$K=2$
        & 1893.4 $\pm$ 5.9
        & 0.8322
        & 0.7808 \\

        \bottomrule
    \end{tabular}
    }
\end{minipage}
\hfill
\begin{minipage}[t]{0.43\textwidth}
    \centering
    \captionof{table}{
    \textbf{Comparison under cross-generator transfer.}
    Wan2.2~\citep{wan2025wan} is replaced with
    CogVideoX~\citep{yang2025cogvideox} at test time
    without further optimization. \textbf{Bold} and \underline{underline} denote the best and runner-up results, respectively.
    }
    \label{tab:transfer}
    \small
    \resizebox{\linewidth}{!}{
    \begin{tabular}{lccc}
        \toprule
        Method
        & ViCLIP V-V $\uparrow$
        & DINOv2 $\uparrow$
        & CLIP-I $\uparrow$ \\
        \midrule

        \multicolumn{4}{l}{Wan2.2 $\rightarrow$ CogVideoX} \\

        FLF-Only
        & 0.7882 & 0.6589 & 0.8297 \\

        PEZ
        & 0.8047 & 0.6318 & 0.8206 \\

        Captioning
        & \underline{0.8222}
        & \underline{0.6615}
        & \underline{0.8307} \\

        \rowcolor{blue!8}
        \textbf{\methodname{} (Ours)}
        & \textbf{0.8709}
        & \textbf{0.7044}
        & \textbf{0.8517} \\

        \bottomrule
    \end{tabular}
    }
\end{minipage}

\end{table*}

\paragraph{\yr{Stronger captioning baselines}}
\yr{To evaluate \methodname{} against stronger captioning baselines, we conduct experiments on MSR-VTT~\citep{xu2016msr}, replacing Qwen2.5-VL-7B~\citep{bai2025qwen25vl} with Qwen3.5-9B~\citep{qwen3.5} and using a more detailed instruction covering visual and temporal content. As shown in Table~\ref{tab:strong_captioning}, the stronger captioning baseline improves over the original one, while \methodname{} remains the best across all metrics. Based on the prompt variants we explored, more elaborate prompting does not provide a clear improvement over standard Qwen3.5-9B captioning. These results indicate that the improvement of \methodname{} cannot be explained simply by a stronger VLM or more elaborate prompting.}

\begin{table}[t]
    \vspace{-3mm}
    \centering
    \caption{
    \yr{\textbf{Comparison with stronger captioning baselines on MSR-VTT.} 
    \textbf{Bold} and \underline{underline} denote the best and runner-up results, respectively.}
    }
    \label{tab:strong_captioning}
    \vspace{2mm}
    
    \small
    \setlength{\tabcolsep}{6pt}
    \begin{tabular}{lccc}
        \toprule
        Method 
        & ViCLIP V-V $\uparrow$ 
        & DINOv2 $\uparrow$ 
        & CLIP-I $\uparrow$ \\
        \midrule
        Captioning (Qwen2.5-VL-7B)~\citep{bai2025qwen25vl}
        & 0.8021
        & \underline{0.6214}
        & 0.7872 \\

        Captioning (Qwen3.5-9B)~\citep{qwen3.5}
        & \underline{0.8312}
        & 0.6207
        & \underline{0.8063} \\

        \quad + Detailed instruction
        & 0.8123
        & 0.6206
        & 0.7932 \\

        \rowcolor{blue!8}
        \textbf{\methodname{} (Ours)}
        & \textbf{0.8434}
        & \textbf{0.6516}
        & \textbf{0.8091} \\
        \bottomrule
    \end{tabular}
\end{table}









\paragraph{\yr{Cross-generator transfer}}
\yr{We further assess cross-model transfer by replacing the Wan2.2~\citep{wan2025wan} generator used during meta-training with CogVideoX~\citep{yang2025cogvideox} at test time, while keeping the learned meta-prompt fixed. As shown in Table~\ref{tab:transfer}, \methodname{} consistently outperforms the baselines on all three reported metrics, suggesting that the learned meta-prompt is not tightly coupled to a specific video generator. Additional implementation details are provided in Appendix~\ref{appen:implementation_details}.
}





\subsection{Downstream Application: Prompt-Based Video Editing}
\label{sec:downstream_editing}

\yr{To evaluate the effectiveness of \methodname{} in downstream video editing, we consider the subject-motion editing tasks of IVEBench~\citep{chen2026ivebench} as a representative downstream editing benchmark. We follow its evaluation protocol and compare \methodname{} with two video editing methods: RF-Edit~\citep{wang2025taming}, an inversion-based editing framework for rectified-flow models, and Wan-Edit~\citep{li2025five}, a training-free video editing method for the Wan model family. For \methodname{}, we use GPT-4o~\citep{hurst2024gpt} to minimally rewrite the recovered inverse prompt according to the edit instruction and check whether the requested motion is reflected in the edited prompt. The edited prompt and only the first frame of the source video are then used for video generation, without access to the benchmark target prompt or any editing-specific training.}

\paragraph{Quantitative results}
\yr{As shown in Table~\ref{tab:editing_results}, despite requiring no editing-specific training, the \methodname{}-based prompt editing pipeline achieves the highest Instruction Satisfaction and Content Fidelity while remaining competitive across the other metrics. In particular, \methodname{} remains competitive with RF-Edit~\citep{wang2025taming}, which is built upon a substantially larger video model with an RF solver, and Wan-Edit~\citep{li2025five}, which is specifically designed for editing within the same model family. These results show that inverse prompts recovered for reconstruction can also serve as effective control interfaces for downstream video editing. Further details on the evaluation metrics and additional quantitative results are provided in Appendix~\ref{appen:downstream_editing}}.

\paragraph{Qualitative results}
\yr{Figure~\ref{fig:editing_results} further illustrates that simple textual modification of the recovered inverse prompt can induce the intended motion changes in the generated video. For the instruction ``Make the man and the boy suddenly fall asleep,'' \methodname{} progressively changes their postures, with the man reclining backward and the boy lowering his head, while preserving the source scene and overall composition. Compared with \methodname{}, RF-Edit~\citep{wang2025taming} shows greater changes in subject appearance and scene content in later frames, while Wan-Edit~\citep{li2025five} reflects the requested sleeping motion less clearly. Table~\ref{tab:editing_prompt_example} further shows how the recovered inverse prompt is modified according to the edit instruction. In this example, the action of examining an object is changed to waving the other hand, while the surrounding scene details are preserved. This illustrates how the recovered prompt can support targeted motion editing while retaining the original scene context.
}

\begin{table*}[t]
    \centering
    \caption{
        \yr{
        \textbf{Quantitative comparison of video editing performance.}
        We compare \methodname{} with RF-Edit~\citep{wang2025taming} and Wan-Edit~\citep{li2025five} under the same edit instructions
        in terms of instruction compliance, video quality, and content fidelity.
        Best results are marked in \textbf{bold}.
        }
    }
    \label{tab:editing_results}

    \resizebox{0.85\linewidth}{!}{
    \begin{tabular}{llccc}
        \toprule
        \textbf{Dimension} & \textbf{Metric} & \textbf{RF-Edit} & \textbf{Wan-Edit} & \textbf{\methodname{} (Ours)} \\
        \midrule

        \multirow{3}{*}{Compliance}
        & Instruction Satisfaction          & 3.09 & 2.09 & \textbf{3.25} \\
        & Phrase Semantic Consistency       & 0.18 & \textbf{0.19} & \textbf{0.19} \\
        & Overall Semantic Consistency      & 0.23 & \textbf{0.25} & \textbf{0.25} \\
        \midrule

        \multirow{5}{*}{Quality}
        & Subject Consistency               & 0.90 & \textbf{0.93} & 0.92 \\
        & Background Consistency            & 0.94 & \textbf{0.95} & \textbf{0.95} \\
        & Temporal Flickering               & \textbf{0.98} & 0.97 & 0.97 \\
        & Motion Smoothness                 & \textbf{0.99} & 0.98 & \textbf{0.99} \\
        & Video Training Suitability Score  & 0.02 & \textbf{0.05} & \textbf{0.05} \\
        \midrule

        \multirow{2}{*}{Fidelity}
        & Content Fidelity                  & 3.06 & 3.81 & \textbf{3.95} \\
        & Semantic Fidelity                 & 0.79 & \textbf{0.97} & 0.96 \\
        \bottomrule
    \end{tabular}
    }
\end{table*}

\begin{figure*}[t]
    \centering
    \includegraphics[
        width=\textwidth,
        trim={0 90 0 90},
        clip
    ]{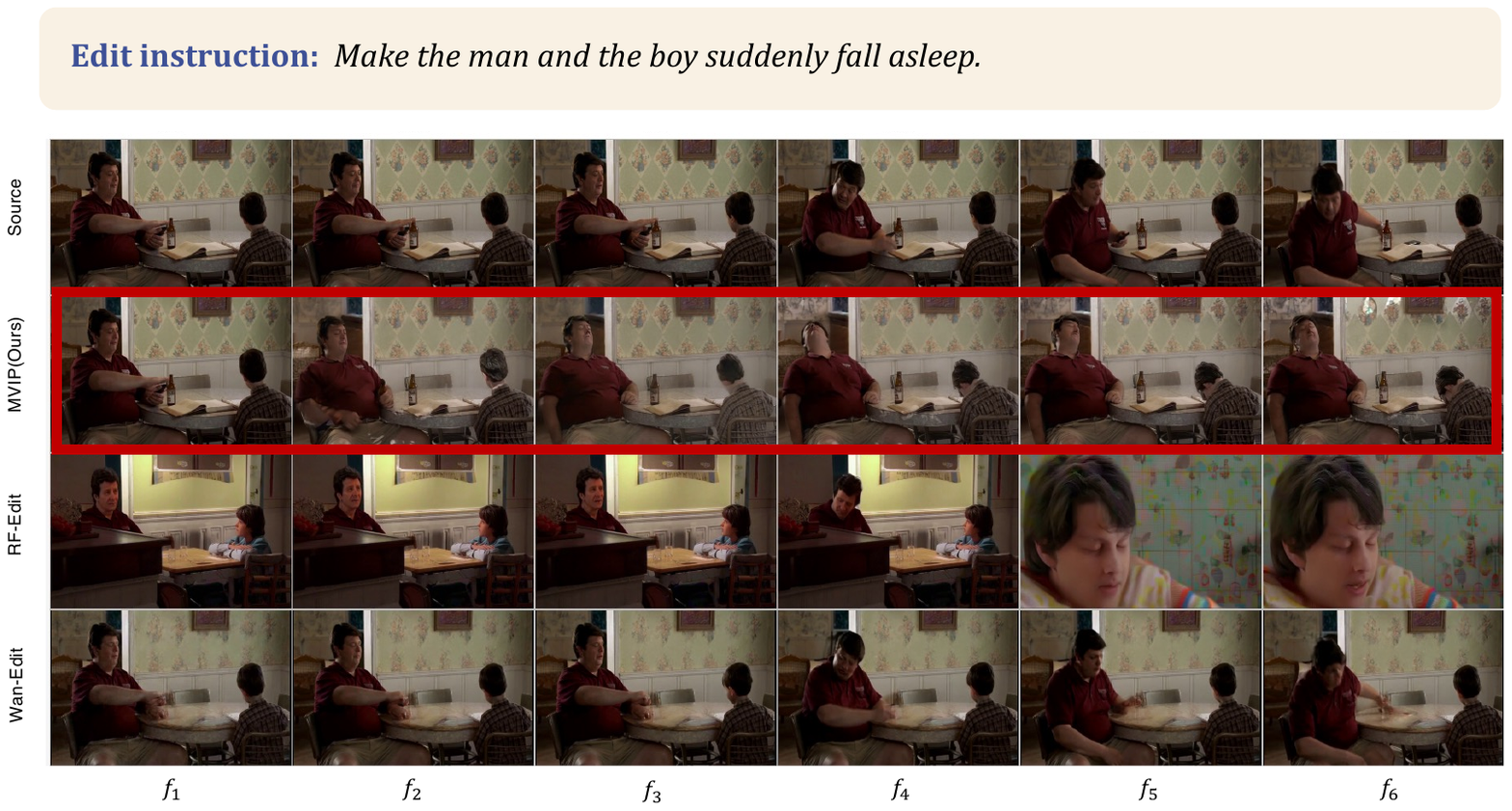}
    \vspace{-3mm}
    \caption{
        \yr{
        \textbf{Qualitative comparison on prompt-based video editing.}
        All methods use the same source video and edit instruction. More qualitative results can be found in Appendix~\ref{appen:downstream_editing}.
        }
    }
    \label{fig:editing_results}
\end{figure*}

\begin{table}[t]
    \centering
    \caption{
        \yr{
        \textbf{Prompt transformations for downstream video editing.}
        For an IVEBench~\citep{chen2026ivebench} example, we show the prompts recovered and edited by \methodname{}, with the benchmark's source and target prompts for reference.
        Phrases modified to reflect the edit instruction are highlighted in \textbf{bold}.
        }
    }
    \label{tab:editing_prompt_example}
    \vspace{2mm}
    
    \small

    \begin{tabularx}{\textwidth}{
        @{}
        p{0.14\textwidth}
        >{\raggedright\arraybackslash}X
        >{\raggedright\arraybackslash}X
        @{}
    }
        \toprule

        \textbf{Method}
        &
        \hfill
        \shortstack{
            \textbf{Before editing}\\
            \scriptsize Source / Recovered
        }
        \hfill\mbox{}
        &
        \hfill
        \shortstack{
            \textbf{After editing}\\
            \scriptsize Target / Edited
        }
        \hfill\mbox{}
        \\

        \midrule

        \textbf{IVEBench}
        &
        A woman is enjoying a sunny day on a yacht, smiling.
        &
        A woman is \textbf{waving her hand} and enjoying
        a sunny day on a yacht, smiling.
        \\[3pt]

        \textbf{MVIP (Ours)}
        &
        The, the woman is standing on a boat,
        holding a small object in her hand and
        \textbf{examining it closely}.
        She appears to be focused, \ldots
        &
        The, the woman is standing on a boat,
        holding a small object in one hand and
        \textbf{waving her other hand}.
        She appears to be focused, \ldots   
        \\

        \bottomrule
    \end{tabularx}
\end{table}

\begin{table*}[t]
\centering

\begin{minipage}[t]
{0.55\textwidth}
    \centering
    \vspace{-2mm}
    \captionof{table}{
        \textbf{\yr{Ablation study of \methodname{} on MSR-VTT.}}
        \yr{We compare reconstruction performance with and without the meta-update and Best-of-$N$ candidate selection.
        Best results are marked in \textbf{bold}.}
    }
    \label{tab:main_ablations}

    \resizebox{\linewidth}{!}{
    \begin{tabular}{lccccc}
        \toprule
        Method
        & Meta
        & BoN
        & ViCLIP V-V $\uparrow$
        & DINOv2 $\uparrow$
        & CLIP-I $\uparrow$ \\
        \midrule

        \rowcolor{blue!8}
        \textbf{\methodname{} (Ours)}
        & $\checkmark$ & $\checkmark$
        & \textbf{0.8434}
        & \textbf{0.6516}
        & \textbf{0.8091} \\

        \methodname{} (Ours)
        & $\checkmark$ & $\times$
        & 0.8389 & 0.6174 & 0.8081 \\

        \methodname{} (Ours)
        & $\times$ & $\checkmark$
        & 0.8314 & 0.6303 & 0.8057 \\

        \midrule
        Captioning
        & -- & --
        & 0.8021 & 0.6214 & 0.7872 \\

        \bottomrule
    \end{tabular}
    }
\end{minipage}
\hfill
\begin{minipage}[t]{0.42\textwidth}
    \centering
    \vspace{-2mm}
    \captionof{table}{
        \textbf{Ablation on meta-prompt initialization and configuration.}
        \textbf{Bold} and \underline{underline} denote the best and runner-up results,
        respectively.
    }
    \label{tab:phi_initialization}

    \small
    \setlength{\tabcolsep}{3.5pt}
    \renewcommand{\arraystretch}{1.08}

    \begin{adjustbox}{max width=\linewidth}
    \begin{tabular}{cccccc}
    \toprule
    Prompt init.
    & $|\Phi|$
    & $K$
    & ViCLIP-VV $\uparrow$
    & DreamSim $\downarrow$
    & CLIP-I $\uparrow$ \\
    \midrule

    \rowcolor{blue!8}
    Natural $\Phi$
    & 16
    & 1
    & \textbf{0.9193}
    & \textbf{0.0917}
    & \textbf{0.8993} \\

    \midrule

    Random $\Phi$
    & \textcolor{black!35}{4}
    & \textcolor{black!35}{1}
    & 0.9064
    & 0.0986
    & 0.8926 \\

    \textcolor{black!35}{Natural $\Phi$}
    & 4
    & \textcolor{black!35}{1}
    & 0.9129
    & 0.1010
    & 0.8960 \\

    \textcolor{black!35}{Natural $\Phi$}
    & 8
    & \textcolor{black!35}{1}
    & 0.9045
    & 0.1021
    & 0.8916 \\

    \textcolor{black!35}{Natural $\Phi$}
    & \textcolor{black!35}{16}
    & 3
    & \underline{0.9174}
    & \underline{0.0971}
    & \underline{0.8992} \\

    \bottomrule
\end{tabular}
    \end{adjustbox}
\end{minipage}
\end{table*}

\subsection{Ablation study}
\paragraph{\yr{Effect of meta-learning}}
\yr{To examine whether the improvement comes from the meta-learning paradigm itself or simply from optimization with video-generation feedback, we compare \methodname{} with an ablated variant without the meta-update.
As shown in Table~\ref{tab:main_ablations}, the variant without meta-learning outperforms direct captioning but remains consistently below \methodname{} across all metrics. This indicates that the improvement cannot be explained solely by optimization under the reconstruction objective, but also benefits from learning a task-adaptive initialization through meta-learning.}

\yr{\paragraph{Effect of Best-of-$N$ sampling}
We compare the default Best-of-$N$ setting ($N=4$) with single-candidate sampling ($N=1$) during inner-loop adaptation. As shown in Table~\ref{tab:main_ablations}, using multiple candidate prompts consistently improves all evaluation metrics. This suggests that ranking sampled candidates by their reconstruction scores provides a higher-quality pseudo-target for meta-prompt adaptation under the black-box reconstruction objective.}

\paragraph{Meta-prompt initialization}
Table~\ref{tab:phi_initialization} analyzes meta-prompt initialization by comparing randomly initialized soft tokens and natural soft-token initialization. All configurations are evaluated on a held-out
validation subset. Natural $\Phi$ initializes the learnable soft tokens from the token embeddings of the instruction, ``Describe objects, actions, camera motion, and temporal changes,'' while random $\Phi$ uses Gaussian samples of the same length. The results show that natural $\Phi$ provides a better starting point for meta-learning and achieves stronger reconstruction quality than random $\Phi$ of the same length across most metrics, suggesting that a meaningful textual prior
offers a better initial basis for meta-learning.

\paragraph{Effect of meta-prompt length}
We examine the influence of the number of learnable soft tokens $|\Phi|$.
Too few tokens limit the meta-prompt capacity, while longer prompts bring little additional benefit. As shown in Table~\ref{tab:phi_initialization}, using $|\Phi|=16$ rather
than $|\Phi|=4$ improves ViCLIP V-V from 0.9129 to 0.9193, with $|\Phi|=16$ achieving the best score. Notably, the best and second-best results across metrics are also obtained by configurations with $|\Phi|=16$, showing that this
length is consistently effective. This suggests that a compact but sufficiently
expressive set of learnable tokens provides an effective prompt prior without
requiring a long prompt representation.

\paragraph{Inner-loop adaptation depth}
We also vary the number of inner-loop adaptation steps $K$ during meta-training.
Table~\ref{tab:phi_initialization} shows that increasing $K$ from 1 to 3 does not
translate into better overall performance. Compared with $K=1$, $K=3$ slightly lowers ViCLIP V-V and CLIP-I and
worsens DreamSim. Moreover, as detailed in
Appendix~\ref{appen:results}, using three inner-loop steps takes about
3$\times$ longer per outer update than using one inner-loop step. Since this additional cost does not lead to clear performance gains, we use $K=1$ as our default configuration.

\section{Conclusion}
In this paper, we have presented \methodname{}, a meta-learning framework for recovering inverse prompts from videos. While prior inverse prompting methods focus on
images, we show that video inverse prompting requires a different approach due
to temporal dynamics and the high cost of video optimization. Through comparisons with baselines, we show that \methodname{} improves
video reconstruction using only a compact set of meta-learned soft tokens, without
updating the VLM or the video generator. These results suggest that video inverse
prompting can be effectively amortized through a learned prompt prior, which
provides a reusable initialization for retaining scene context and temporal cues.

\begin{ack}

We thank Yunhun Nam for his help in improving our manuscript.
This work was supported by Institute of Information \& communications Technology Planning \& Evaluation (IITP) grants funded by the Korea government (MSIT) 
(Development of Human-Like Intelligent Generative Agents Based on Interactive Multimodal Reverse Prompting (No.~RS-2026-25519206, 25\%); 
Development of Media Vibe Editing Technology (No.~RS-2026-25519202, 25\%);
Artificial Intelligence Star Fellowship Support Program to Nurture the Best Talents (No.~IITP-2026-RS-2025-02304828, 20\%);
Development of AI Co-Scientist based on Scientific Causal World Model (No.~RS-2026-25507543, 10\%);
Artificial Intelligence Graduate School Program (Korea University) (No.~RS-2019-II190079, 5\%)),
the National Research Foundation of Korea (NRF) grant funded by the Korea government (MSIT) (No.~RS-2025-23523603, 10\%), and
the Korea Creative Content Agency grant funded by the Ministry of Culture, Sports and Tourism (International Collaborative Research and Global Talent Development for the Development of Copyright Management and Protection Technologies for Generative AI (No.~RS-2024-00345025, 5\%)).

We also acknowledge the compute supports by the National Supercomputing Center (KSC-2025-CRE-0511) and the Advanced GPU Utilization Support Program funded by the Korea Government (Ministry of Science and ICT).


\end{ack}

\bibliographystyle{plainnat}
{\small
\bibliography{references}
}

\clearpage
\appendix

\section{Limitations and Broader Impact}
\label{app:limitation_broader_impact}
\paragraph{Limitations}
While our work demonstrates the feasibility of practical video inverse prompting, several aspects remain beyond the scope of the current framework. In particular, the proposed method assumes a first-last-frame (FLF) conditioning setup, where the boundary frames of the target video are provided. This design choice allows the model to focus on learning temporally coherent inverse prompting beyond static appearance reconstruction, and is also motivated by the fact that many recent public video generative models support FLF-conditioned generation. Nevertheless, the current framework does not address fully text-only video inverse prompting, which remains a challenging open problem due to the severe ambiguity of prompt-video mappings in the video domain. Despite this limitation, the proposed approach can still serve as a practical component within broader media generation pipelines when combined with existing image-based inverse prompting and editing methods. In addition, further developments may be required for deployment in real-world media production workflows. For example, learning more structured and editable captioning schemas through meta-learning could improve controllability and downstream editing flexibility, which we view as a promising direction for future work.

\paragraph{Broader impact}
This work can provide positive societal impacts by improving the controllability of video generative models, which benefits applications such as content creation, media production, and simulation. In particular, enabling reconstruction-aware inverse prompting may reduce the manual trial-and-error process currently required for prompt engineering, improving accessibility and efficiency for non-expert users. At the same time, the framework could potentially be misused to facilitate unauthorized reproduction or manipulation of generated media, including malicious editing through video generation pipelines. Furthermore, recovering prompts from videos may raise concerns regarding intellectual property when applied to proprietary or copyrighted generative content. While our work focuses on improving controllable generation workflows rather than bypassing model safeguards or recovering hidden prompts exactly, we acknowledge these potential risks and encourage future research on responsible deployment, access control, and misuse monitoring for inverse prompting systems.

\section{Experimental Details}
\label{appen:implementation_details}

\paragraph{Models}
We use Qwen2.5-VL-7B-Instruct~\citep{bai2025qwen25vl} as the VLM for video captioning and reference--generation comparison.
Two Wan2.2~\citep{wan2025wan} checkpoints are used. During meta-training, the inner-loop reconstruction score in Eq.~\eqref{eq:bestofn_selection} is computed with the first-frame-conditioned \texttt{Wan-AI/Wan2.2-TI2V-5B-Diffusers}. All reported evaluations, including every baseline, use the first-last-frame-conditioned \texttt{alibaba-pai/Wan2.2-Fun-5B-InP}~\citep{aigc_apps_VideoX_Fun_2026}; the meta-learned $\Phi$ is transferred to this checkpoint without any further optimization.
For the cross-generator evaluation in Table~\ref{tab:transfer}, we additionally use \texttt{alibaba-pai/CogVideoX-Fun-V1.5-5b-InP}~\citep{aigc_apps_VideoX_Fun_2026,yang2025cogvideox} at test time only. First-last-frame conditioning for both evaluation generators is provided by VideoX-Fun~\citep{aigc_apps_VideoX_Fun_2026}. In the refinement baseline, prompt rewriting is performed by GPT-4o-mini~\citep{hurst2024gpt}; prompt editing in Section~\ref{sec:downstream_editing} uses GPT-4o~\citep{hurst2024gpt}. All VLM and video generator weights are kept frozen throughout the experiments.

\paragraph{Video generation}
We use 30 inference steps for video generation. 
The length of each generated video is matched to the corresponding reference segment and then adjusted to the nearest valid temporal length of the form $4k+1$, as required by the generator. 
Unless otherwise specified, we use a fixed random seed of 42. 
For seed robustness analysis, we additionally use seeds 43 and 44. 

\paragraph{Segmentation and conditioning}
All videos are split into scene-consistent segments using PySceneDetect with a threshold of 27. 
For each segment, the first and last frames are used as visual conditions for the video generator. 
The same segmentation and conditioning protocol is applied to both \methodname{} and all baselines.

\paragraph{Datasets}
\begin{itemize}
    \item \textbf{DAVIS-2017.}
    \yr{For evaluation on unseen datasets, we use all 30 sequences from the
    DAVIS-2017 validation split~\cite{perazzi2016benchmark,pont20172017},
    without additional filtering. Each sequence is treated as a single
    evaluation unit. Video generation uses 30 inference steps, a fixed random
    seed of 42, and a maximum generation length of 121 frames. For prompt
    generation, we uniformly sample 8 frames as input to the VLM, while PEZ
    uses 4 uniformly sampled frames.}

    \item \textbf{Kinetics-400.}
    \yr{We randomly sample 30 clips with seed 42 from the first shard of the
    Kinetics-400 validation split~\cite{kay2017kinetics}, which contains approximately 1,000 videos. We segment each clip using PySceneDetect
    with a threshold of 27 and a minimum scene length of 15 frames.
    Scenes longer than 5 seconds are further divided into shorter segments,
    resulting in 75 evaluation units in total. Video generation follows the
    same settings as for DAVIS-2017.}
\end{itemize}

\paragraph{Meta-learning hyperparameters}
\methodname{} uses a prefix-based soft prompt consisting of 16 learnable soft tokens. 
The soft prompt is initialized from the token embeddings of the instruction 
``Describe objects, actions, camera motion, and temporal changes.'' 
During meta-training, we use Best-of-$N$ prompt sampling with $N=4$, an inner learning rate of $\alpha=10^{-3}$, a meta-step size of $\eta=0.1$, and 100 outer-loop optimization steps.

\paragraph{Compute infrastructure}
Our experiments are conducted on compute cluster instances equipped with a single NVIDIA B200 GPU, each with 183GB of VRAM. 

\section{Evaluation Metrics}
\label{appen:metrics}

Let $V$ and $\hat{V}$ denote the reference video segment and the generated video, respectively.

\paragraph{Frame-level metrics}
We evaluate frame-level visual correspondence between the reference and generated videos. We uniformly sample $T=8$ frames from each video over its temporal extent and pair frames sampled at corresponding relative temporal positions, yielding $\{(v_t,\hat{v}_t)\}_{t=1}^{T}$.

\begin{itemize}
    \item \textbf{CLIP-I}
    We compute the cosine similarity between CLIP~\citep{radford2021learning} image features of corresponding sampled frames and average over time:
    \[
    S_{\mathrm{CLIP\text{-}I}}
    =
    \frac{1}{T}
    \sum_{t=1}^{T}
    \cos\left(
    \mathbf{E}_{\mathrm{CLIP}}(v_t),
    \mathbf{E}_{\mathrm{CLIP}}(\hat{v}_t)
    \right).
    \]

    \item \textbf{DINOv2}
    We compute the cosine similarity between DINOv2~\citep{oquab2024dinov} features of corresponding sampled frames and average over time:
    \[
    S_{\mathrm{DINOv2}}
    =
    \frac{1}{T}
    \sum_{t=1}^{T}
    \cos\left(
    \mathbf{E}_{\mathrm{DINOv2}}(v_t),
    \mathbf{E}_{\mathrm{DINOv2}}(\hat{v}_t)
    \right).
    \]
    
    \item \textbf{DreamSim}
    We compute the DreamSim~\citep{fu2023dreamsim} perceptual distance between corresponding sampled frames and average over time:
    \[
    D_{\mathrm{DreamSim}}
    =
    \frac{1}{T}
    \sum_{t=1}^{T}
    d_{\mathrm{DreamSim}}(v_t,\hat{v}_t).
    \]
\end{itemize}

\paragraph{Video-level metrics}
We evaluate video-level semantic and temporal alignment using video-level representations.

\begin{itemize}
    \item \textbf{ViCLIP V-V}
    We encode 8 uniformly sampled frames from each video using ViCLIP~\citep{wang2024internvid} and compute their cosine similarity:
    \[
    S_{\mathrm{ViCLIP}}
    =
    \cos\left(
    \mathbf{E}_{\mathrm{ViCLIP}}(V),
    \mathbf{E}_{\mathrm{ViCLIP}}(\hat{V})
    \right).
    \]
    
    \item \textbf{VideoMAE}
    We encode 16 uniformly sampled frames from each video using VideoMAE~\citep{tong2022videomae} and compute their cosine similarity:
    \[
    S_{\mathrm{VideoMAE}}
    =
    \cos\left(
    \mathbf{E}_{\mathrm{VideoMAE}}(V),
    \mathbf{E}_{\mathrm{VideoMAE}}(\hat{V})
    \right).
    \]

    \item \textbf{RAFT Dist.}
    We estimate optical flow for corresponding adjacent frame pairs using RAFT~\citep{teed2020raft} and measure the differences in flow magnitude and direction. The final distance is the equally weighted average of the magnitude and direction distances:
    \[
    D_{\mathrm{RAFT}}
    =
    \frac{1}{2}D_{\mathrm{mag}}
    +
    \frac{1}{2}D_{\mathrm{dir}}.
    \]
\end{itemize}

\paragraph{Text-alignment metrics}
We additionally evaluate the semantic alignment between the generated text prompts and the target videos.
\begin{itemize}
    \item \textbf{ViCLIP V-T} We compute the cosine similarity between the ViCLIP~\citep{wang2024internvid} video embedding of the target video and the text embedding of the generated prompt.
    \item \textbf{SigLIP} We compute the SigLIP~\citep{zhai2023sigmoid} image-text similarity between each sampled frame of the target video and the generated prompt, averaged over frames.
    \item \textbf{BLIP2-ITM} We compute the image-text matching score of BLIP-2~\citep{li2023blip} between each sampled frame of the target video and the generated prompt, averaged over frames.
\end{itemize}

\paragraph{VBench metrics}
We evaluate general video generation quality using the official VBench \cite{huang2024vbench} evaluation pipeline.

\begin{itemize} 
\item \textbf{Motion smoothness} Quantifies the temporal continuity and physical plausibility of the generated motion. This metric evaluates whether object movements and camera motions evolve smoothly across consecutive frames, without sudden jumps, discontinuous trajectories, or unnatural transitions. A higher score indicates that the motion follows a coherent temporal progression and appears more realistic. 
\item \textbf{Dynamic degree} Measures the amount of motion contained in the generated video. This metric evaluates whether the model produces sufficiently dynamic content when motion is expected, rather than generating nearly static frames. A higher score indicates that the generated video contains more noticeable object movement, camera motion, or scene-level dynamics. 
\item \textbf{Subject consistency} Measures the temporal consistency of the main subject throughout the video. This metric evaluates whether the subject preserves its identity, shape, appearance, and visual attributes across frames, even when it moves or changes position. A higher score indicates that the generated subject remains recognizable and does not suffer from identity drift, deformation, or unintended appearance changes. 
\item \textbf{Background consistency} Measures the stability and temporal coherence of the background scene. This metric evaluates whether the background maintains a consistent layout, structure, and visual appearance across frames, instead of changing abruptly or being distorted over time. A higher score indicates that the generated scene remains stable and coherent throughout the video. 
\item \textbf{Temporal flickering} Measures undesirable frame-to-frame visual fluctuations in the generated video. This metric focuses on local and high-frequency temporal artifacts such as sudden brightness changes, texture instability, color shifts, or pixel-level flickering. A higher score indicates less temporal flickering and, therefore, better visual stability across consecutive frames. 

\end{itemize}

\paragraph{Internal optimization scores}
We use internal reconstruction scores during prompt optimization to guide prompt refinement.

\begin{itemize}
    \item \textbf{CLIP score.}
    We compute the frame-level CLIP similarity between the generated and reference videos:
    \[
    S_{\mathrm{CLIP}}
    =
    \frac{1}{T}
    \sum_{t=1}^{T}
    \cos\left(
    \mathbf{E}_{\mathrm{CLIP}}(v_t),
    \mathbf{E}_{\mathrm{CLIP}}(\hat{v}_t)
    \right).
    \]
\end{itemize}      

\section{Baselines}
\label{appen:baselines}

All baselines are evaluated on the same segmented video set as \methodname{}. For each segment, we use the same first-last-frame conditioning, generation seed, video length cap, resolution, and inference budget. Generated videos are evaluated against the same reference segments using the same metric pipeline.

\begin{itemize}
    \item \textbf{FLF-Only.} This baseline evaluates how much reconstruction can be achieved from the video conditioning signal alone, without using any text description of the target video. For each video segment, we use the same first--last frame conditioning as \methodname{}, where the first and last frames of the segment are provided to the video generator. The text prompt is set to an empty string.
    The video is then generated using the same T2V model, resolution, length,
    and inference budget as \methodname{}. This baseline isolates the contribution of
    first--last frame conditioning from the contribution of inverse prompt
    recovery.
    \item \textbf{Captioning.}
    We next consider a direct describe-then-generate pipeline, which provides a
    straightforward text prompt for each target video segment. For each segment, we
    feed the target video to Qwen2.5-VL and ask the model to describe the visible
    scene, subjects, actions, and temporal progression in natural language. The
    generated caption is then used as the text prompt for the video generator,
    together with the same first--last frame conditioning as \methodname{}. No prompt
    optimization, iterative rewriting, or learnable soft prompts are used.
    \item \textbf{PEZ.} PEZ~\cite{wen2023hard} is originally proposed as an image inverse prompting method, while our task requires video-level reverse prompting. Since there is no directly comparable video inverse prompting baseline, we adapt PEZ to the video setting as follows. For each video segment, we uniformly sample four frames across the temporal span of the segment. We then apply PEZ independently to each sampled frame to recover an image-level prompt. The recovered prompts are concatenated in temporal order and used as the text condition for video generation.
    
    \item \textbf{ARPO-based refinement.} ARPO~\cite{ren2025reverse} proposes an iterative feedback-and-rewrite strategy for image reverse prompting, where a prompt is refined by comparing the generated image with the reference image and using the resulting feedback to update the prompt. To adapt this idea to video reverse prompting, we introduce a Captioning+Refine baseline. Starting from a VLM-generated caption using the instruction ``Describe this video in detail,'' we generate an initial reconstruction video. We then uniformly sample frames from both the target video segment and the generated video, and ask Qwen2.5-VL-7B-Instruct to describe their concrete visual and motion differences. The resulting feedback is passed to GPT-4o-mini, which rewrites the current video prompt while preserving its correct content and structure. The refined prompt is then used to regenerate the video under the same first-last-frame conditioning and generation protocol as \methodname{}. To analyze the test-time refinement behavior of a caption-based baseline, we apply this refinement procedure only to the Captioning baseline and compare its
    iteration-wise performance with \methodname{}.
\end{itemize}

\clearpage

\section{Overall Procedure: \methodname{}}

Algorithm \ref{alg:bestofn_reptile} summarizes the full meta-training procedure of \methodname{}. At each outer iteration, we sample a batch of video tasks and
perform task-specific inner-loop adaptation using Best-of-$N$ prompt sampling and reconstruction-based selection. The resulting task-adapted soft prompts are then aggregated through a Reptile-style update, yielding a meta-learned initialization that can be reused for video inverse prompting.

\newcommand{\algcomment}[1]{%
  \hfill{\footnotesize\textcolor{algcommentgray}{\texttt{//}~#1}}%
}

\begin{algorithm}[h]
\caption{\methodname{} Meta-Training}
\label{alg:bestofn_reptile}
\small
\begin{algorithmic}[1]

\STATE \textbf{Input:} task distribution $p(\tau)$ over video segments,
frozen VLM $M_{\psi}$, frozen video generator $G_{\theta}$,
reconstruction score $\mathcal{S}$, sampled-frame index set $\mathcal{R}$,
fixed instruction $h$, inner learning rate $\alpha$,
meta learning rate $\eta$, inner steps $K$,
number of candidates $N$, batch size $B$, outer iterations $T$

\STATE \textbf{Output:} meta-learned soft prompt $\Phi_T$

\STATE \textbf{Initialize:} shared soft prompt $\Phi_0$

\FOR{$t=0,\ldots,T-1$}
    \STATE $\{\tau_b\}_{b=1}^{B}\sim p(\tau)$
    \algcomment{sample video tasks}

    \FOR{$b=1,\ldots,B$}
        \STATE $x_b \gets \{f_r^{v_{\tau_b}}\}_{r\in\mathcal{R}}$
        \algcomment{sample VLM input frames}

        \STATE $v_{b,\mathrm{cond}}
        \gets
        (f_1^{v_{\tau_b}},f_T^{v_{\tau_b}})$
        \algcomment{first-last-frame condition}

        \STATE $\Phi_b^{(0)} \gets \Phi_t$
        \algcomment{initialize task prompt}

        \FOR{$k=1,\ldots,K$}
            \FOR{$i=1,\ldots,N$}
                \STATE $c_{b,i}^{(k)}
                \sim
                p_{\psi}
                (\cdot \mid x_b,h,\Phi_b^{(k-1)})$
                \algcomment{sample candidate prompt}

                \STATE $\hat{v}_{b,i}^{(k)}
                \gets
                G_{\theta}
                (c_{b,i}^{(k)},v_{b,\mathrm{cond}})$
                \algcomment{generate reconstruction}

                \STATE $s_{b,i}^{(k)}
                \gets
                \mathcal{S}
                (\hat{v}_{b,i}^{(k)},v_{\tau_b})$
                \algcomment{score reconstruction}
            \ENDFOR

            \STATE $i^*
            \gets
            \arg\max_{i=1,\ldots,N}
            s_{b,i}^{(k)}$
            \algcomment{best-of-$N$ selection}

            \STATE $c_b^{*,(k)}
            \gets
            c_{b,i^*}^{(k)}$
            \algcomment{reconstruction-guided pseudo-target}

            \STATE $g_b^{(k)}
            \gets
            \nabla_{\Phi_b^{(k-1)}}
            \log
            p_{\psi}
            (c_b^{*,(k)}
            \mid x_b,h,\Phi_b^{(k-1)})$
            \algcomment{VLM likelihood gradient}

            \STATE $\Phi_b^{(k)}
            \gets
            \Phi_b^{(k-1)}
            +\alpha g_b^{(k)}$
            \algcomment{inner soft-prompt update}
        \ENDFOR

        \STATE $\tilde{\Phi}_{t,b}
        \gets
        \Phi_b^{(K)}
        =
        A_{\tau_b}^{K}(\Phi_t)$
        \algcomment{task-adapted soft prompt}
    \ENDFOR

    \STATE $\Phi_{t+1}
    \gets
    \Phi_t+
    \eta\frac{1}{B}
    \sum_{b=1}^{B}
    (\tilde{\Phi}_{t,b}-\Phi_t)$
    \algcomment{Reptile-style meta-update}
\ENDFOR

\STATE \textbf{return} $\Phi_T$
\algcomment{meta-learned soft-prompt initialization}

\end{algorithmic}
\end{algorithm}

\clearpage
\section{Additional Results}
\label{appen:results}
\subsection{Qualitative comparison}
\label{appen:recon_qual}
Figure~\ref{fig:appendix_qual} shows qualitative reconstruction examples. While the baselines capture parts of the target video, they often drift in subject appearance, scene layout, or temporal progression across time. In contrast, \methodname{} preserves the visual content and temporal evolution of the reference video more consistently, especially at later frames. This suggests that the recovered inverse prompt provides stronger reconstruction cues than
FLF-only conditioning, PEZ \cite{wen2023hard}, or direct captioning.

\begingroup
\raggedbottom

\begin{figure}[H]
    \centering

    \begin{subfigure}{\textwidth}
        \centering
        \includegraphics[
            width=\textwidth,
            trim={0 110 0 100},
            clip
        ]{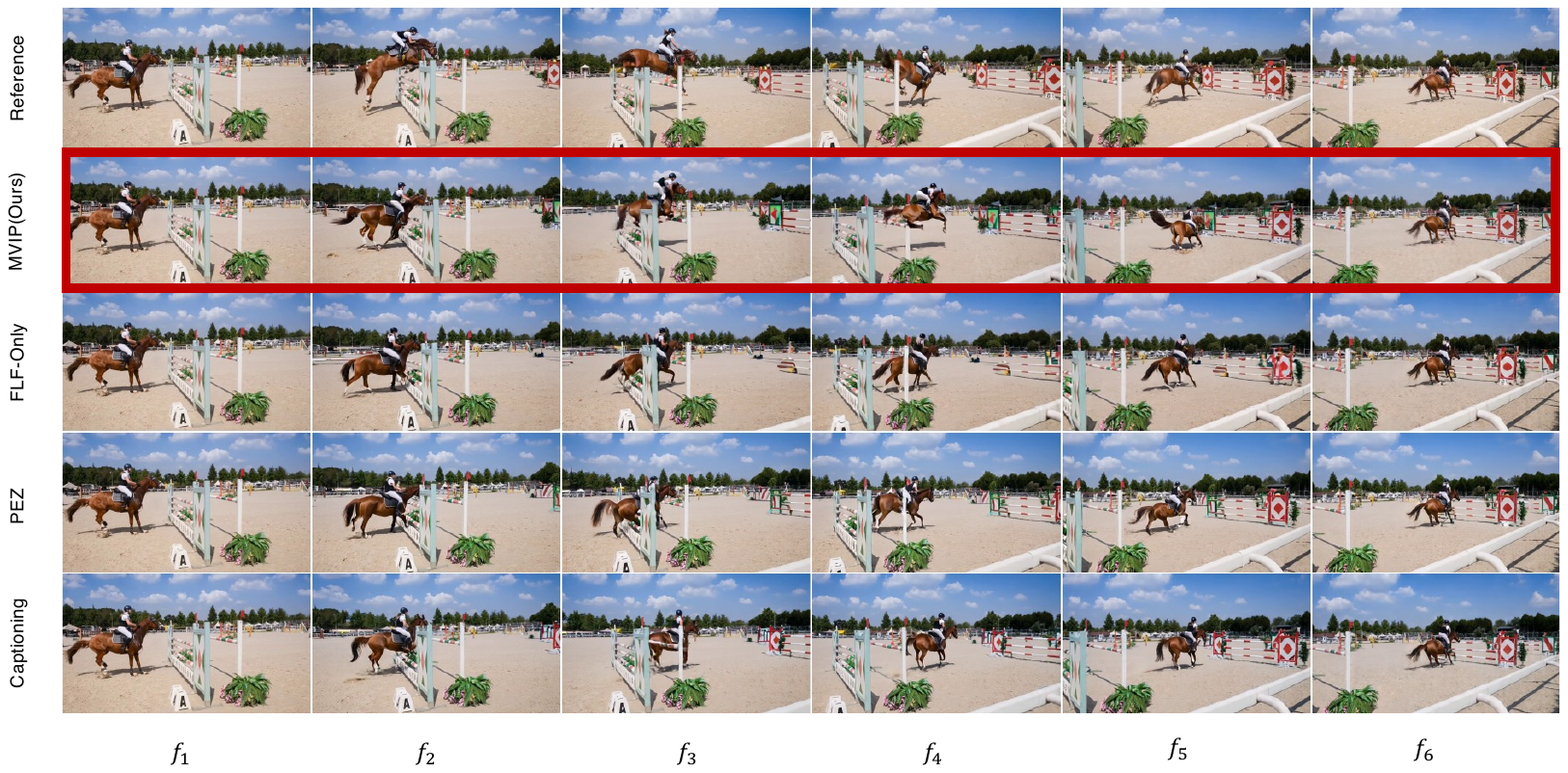}
        \caption{}
        \label{fig:appendix_qual_a}
    \end{subfigure}

    \begin{subfigure}{\textwidth}
        \centering
        \includegraphics[
            width=\textwidth,
            trim={0 110 0 100},
            clip
        ]{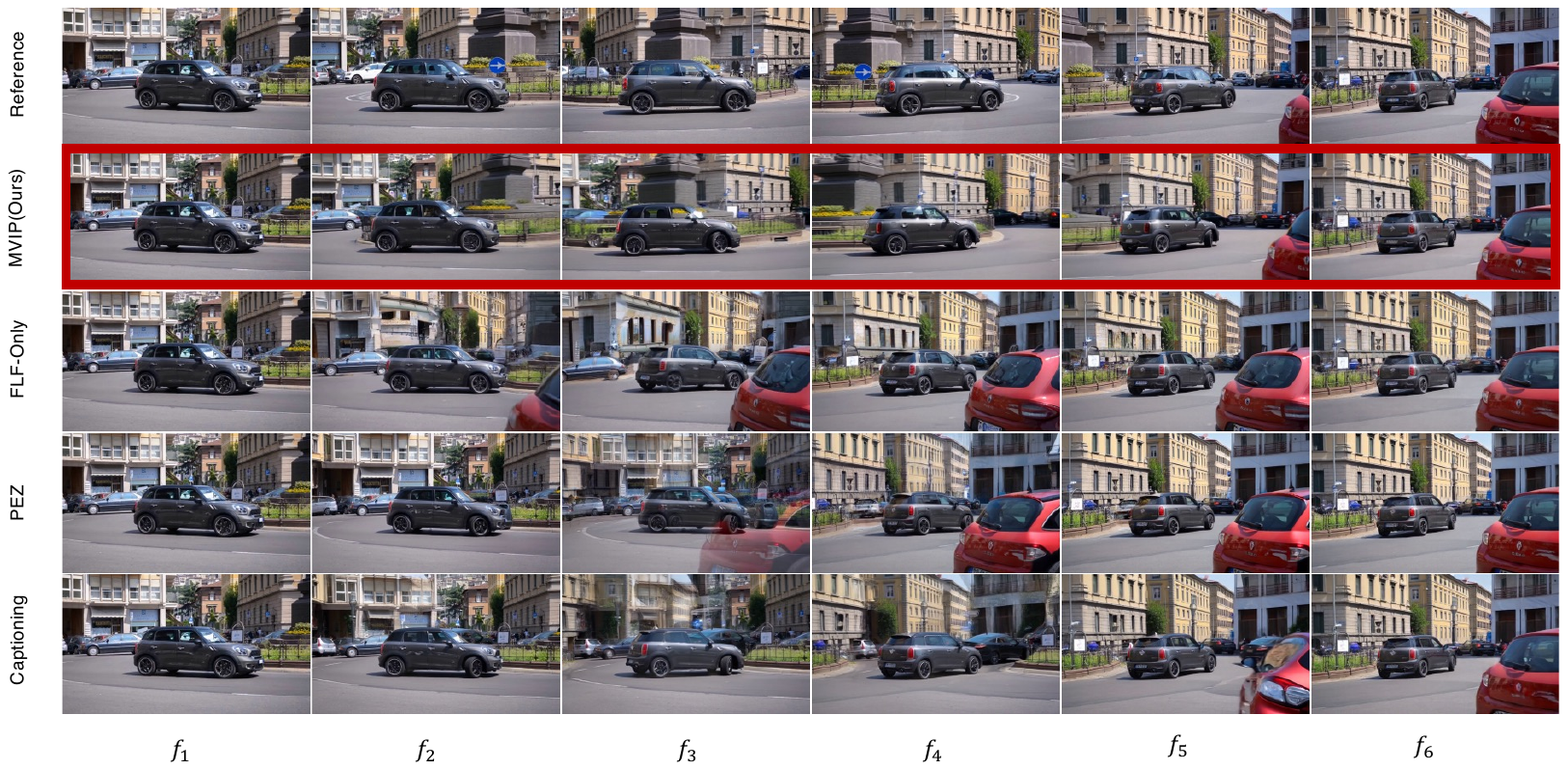}
        \caption{}
        \label{fig:appendix_qual_b}
    \end{subfigure}

    \caption{\textbf{Additional qualitative reconstruction results.}}
    \label{fig:appendix_qual}
\end{figure}

\begin{figure}[H]
    \ContinuedFloat
    \centering

    \begin{subfigure}{\textwidth}
        \centering
        \includegraphics[
            width=\textwidth,
            trim={0 110 0 100},
            clip
        ]{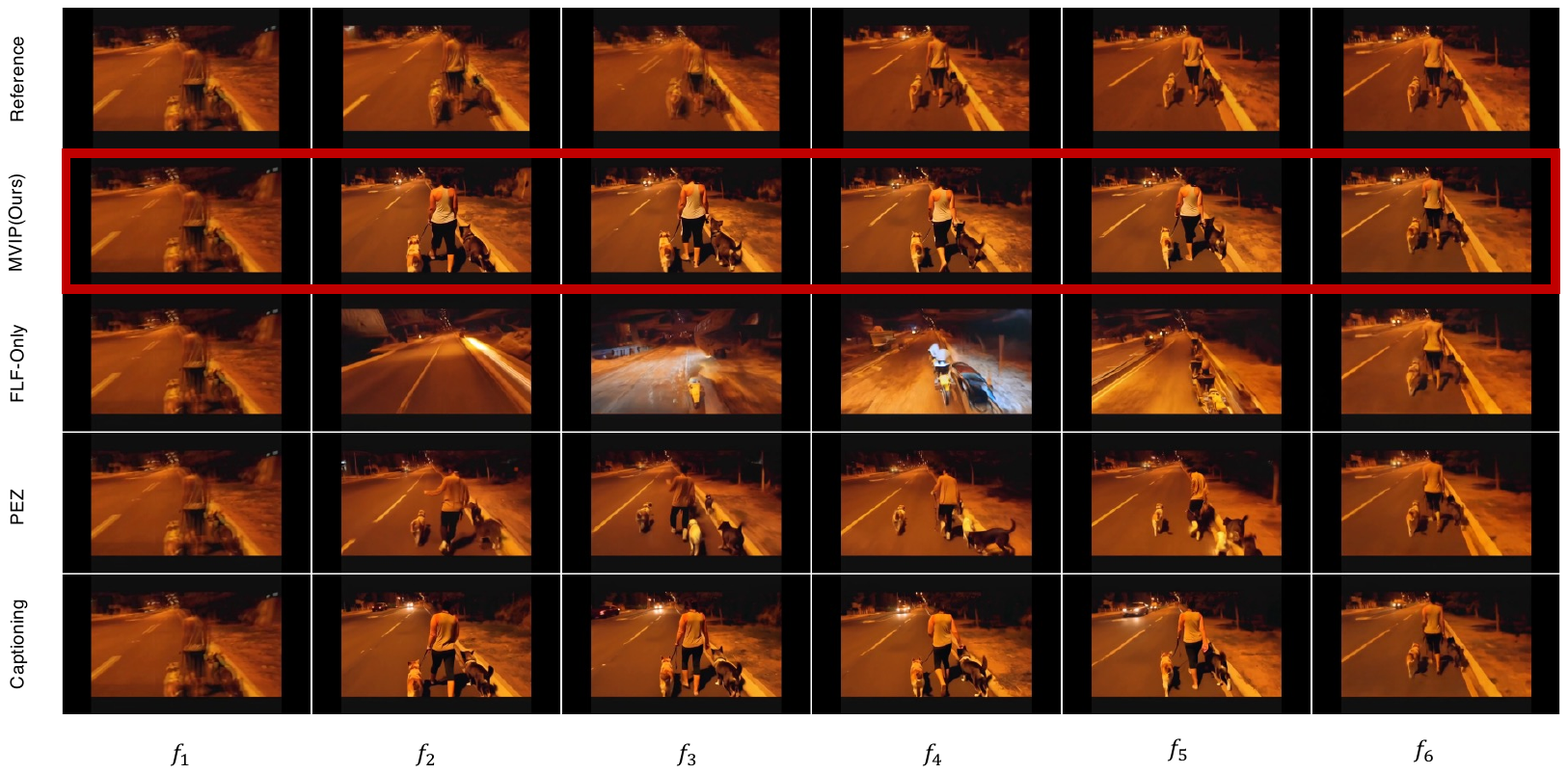}
        \caption{}
        \label{fig:appendix_qual_c}
    \end{subfigure}

    \begin{subfigure}{\textwidth}
        \centering
        \includegraphics[
            width=\textwidth,
            trim={0 110 0 100},
            clip
        ]{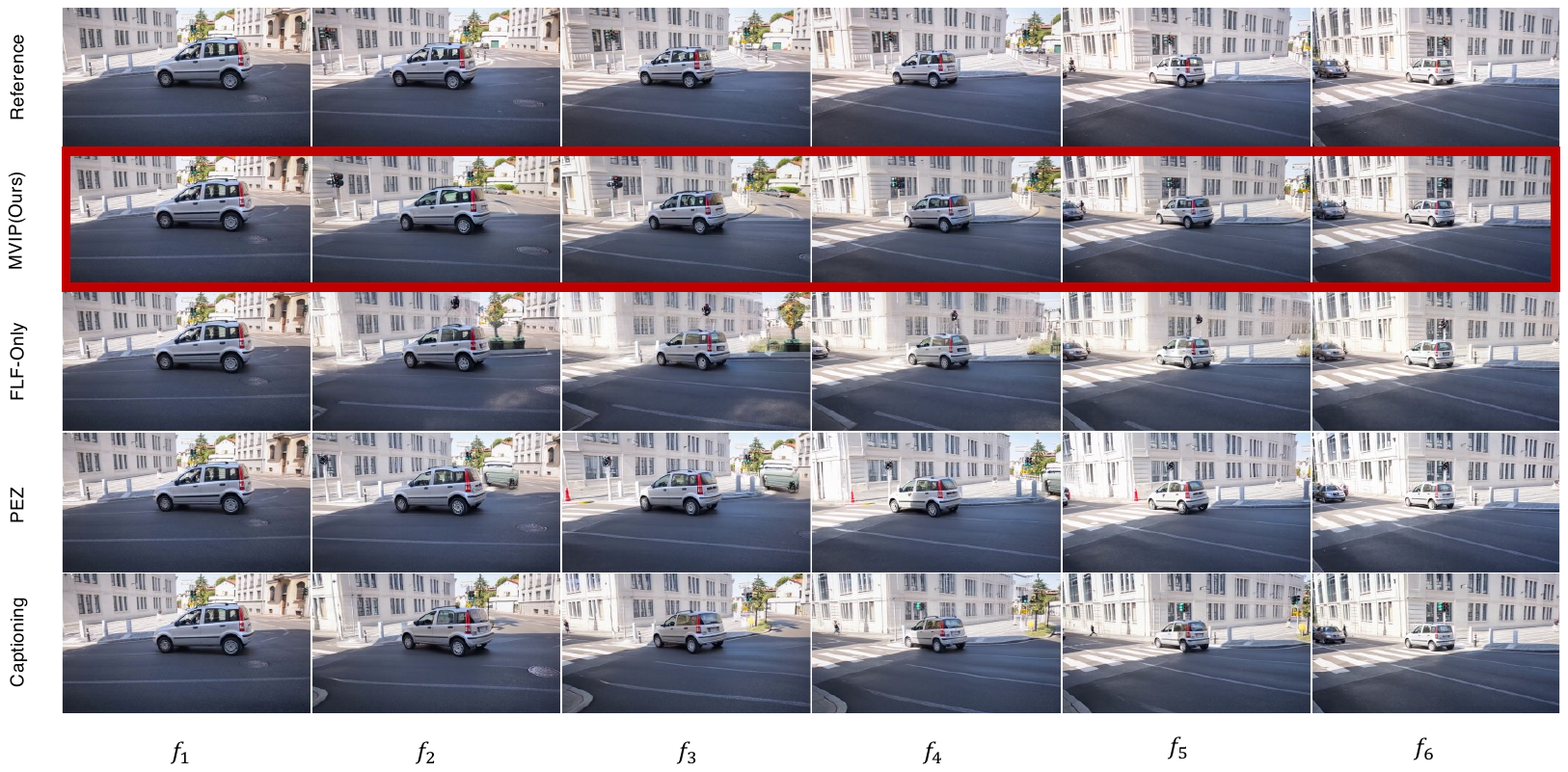}
        \caption{}
        \label{fig:appendix_qual_d}
    \end{subfigure}

    \begin{subfigure}{\textwidth}
        \centering
        \includegraphics[
            width=\textwidth,
            trim={0 110 0 100},
            clip
        ]{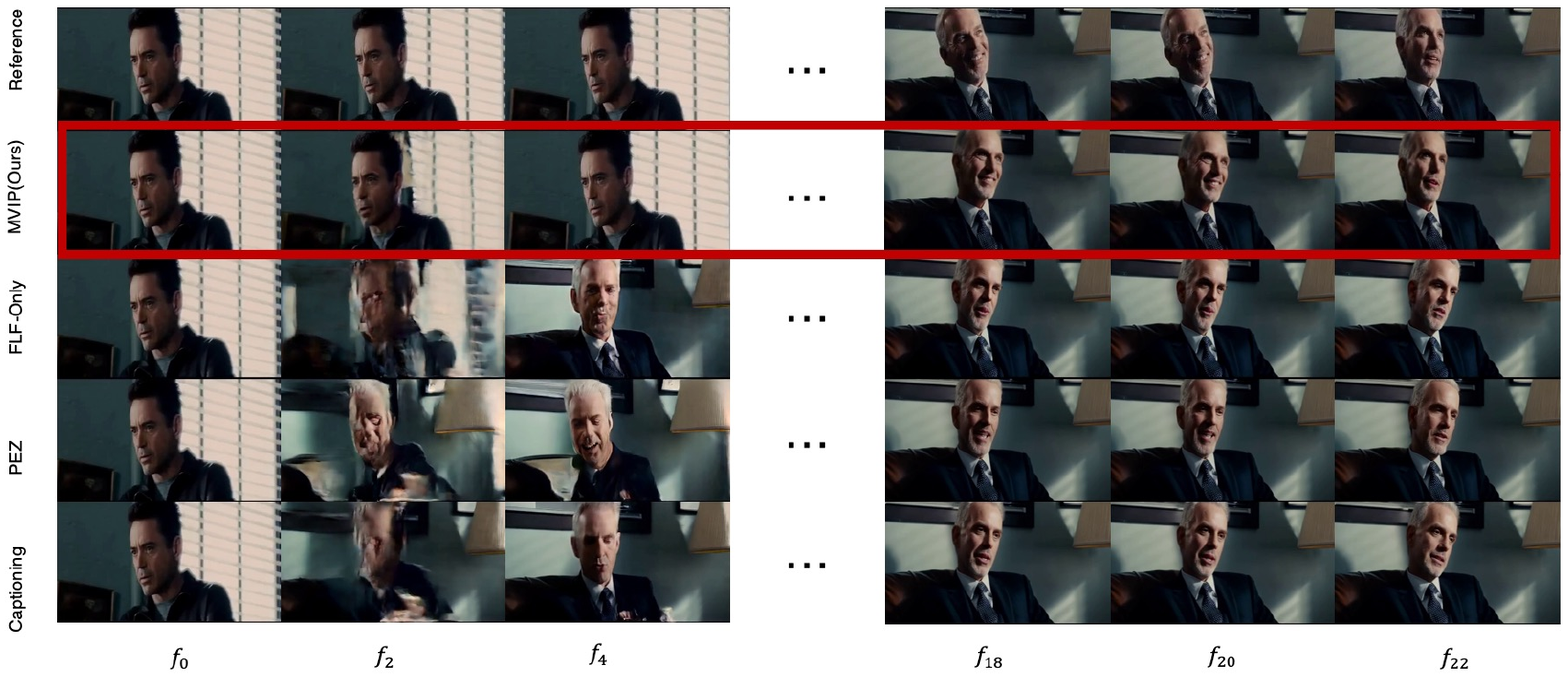}
        \caption{}
        \label{fig:appendix_qual_e}
    \end{subfigure}

    \caption{\textbf{Additional qualitative reconstruction results.}}
\end{figure}


\subsection{Quantitative comparison}
\label{appen:quan}

\paragraph{Iterative caption refinement baseline}
We further compare \methodname{} with an iterative caption refinement baseline adapted from ARPO~\cite{ren2025reverse}. ARPO refines an image prompt by comparing the target image with the generated image and using the observed discrepancies as feedback for the next prompt. To extend this idea to videos, we sample multiple frames from both the target video and the generated video at matched temporal positions.
A VLM is then asked to compare the two sets of frames and identify missing or mismatched visual details, including objects, scene layout, background appearance, motion cues, and temporal changes. The captioning baseline is rewritten using this feedback, and the updated caption is used to generate the next video. We repeat this process for three refinement iterations. Results are reported on the full evaluation set of Table~\ref{tab:main_table}, aggregated over the synthetic and real-world splits.
Table~\ref{tab:caption_refine_iterations} shows that ARPO-style refinement improves the captioning baseline from iteration 0 to iteration 3 on most reported metrics.
However, even after three refinement iterations, the refined baseline remains below \methodname{} at iteration 0 across all metrics. This indicates that test-time caption refinement helps the captioning baseline, but is not sufficient to match the reconstruction quality of our meta-learned inverse prompt.

\begin{table}[H]
\centering
\caption{
\textbf{Comparison with iterative caption refinement.}
We extend the ARPO-style iterative refinement strategy to the video setting by refining the captioning baseline for three iterations.
Although refinement improves the captioning baseline, it remains below MVIP across all reported reconstruction metrics.
}
\label{tab:caption_refine_iterations}

\setlength{\tabcolsep}{4pt}
\begin{tabular}{lcccccc}
\toprule
Method & It. & CLIP-I $\uparrow$ & DINOv2 $\uparrow$ & VideoMAE $\uparrow$ & DreamSim $\downarrow$ & RAFT $\downarrow$ \\
\midrule
Captioning & 0 & 0.858 & 0.733 & 0.969 & 0.191 & 0.471 \\
Captioning & 1 & 0.862 & 0.736 & 0.972 & 0.179 & 0.460 \\
Captioning & 2 & 0.862 & 0.735 & 0.972 & 0.180 & 0.458 \\
Captioning & 3 & 0.861 & 0.733 & 0.972 & 0.180 & 0.459 \\
\rowcolor{gray!12}
Ours & 0 & \textbf{0.870} & \textbf{0.755} & \textbf{0.974} & \textbf{0.168} & \textbf{0.446} \\
\bottomrule
\end{tabular}
\end{table}

\paragraph{\yr{Robustness}}
\yr{We additionally conduct a seed-robustness analysis over three different seeds (viz., 42, 43, and 44) on the synthetic video benchmark used in Table~\ref{tab:main_table}. As shown in Table~\ref{appen:reconstruction_seed_robustness}, the reconstruction performance of \methodname{} remains highly consistent across different random seeds, with small standard deviations on all three metrics, while continuing to outperform the captioning baseline. These results further confirm that the effectiveness of \methodname{} is robust to seed choice rather than depending on the default seed.}

\begin{table}[h]
    \centering
    \small
    \caption{
        \yr{\textbf{Robustness to random seeds on the synthetic video benchmark.}
        Results are reported as mean $\pm$ standard deviation over three
        random seeds (42, 43, and 44).}
    }
    \label{appen:reconstruction_seed_robustness}
    \begin{tabular}{lccc}
        \toprule
        \textbf{Method}
        & \textbf{ViCLIP V-V $\uparrow$}
        & \textbf{DINOv2 $\uparrow$}
        & \textbf{CLIP-I $\uparrow$} \\
        \midrule
        Captioning
        & 0.9281 $\pm$ 0.0018
        & 0.8448 $\pm$ 0.0052
        & 0.9224 $\pm$ 0.0008 \\

        \textbf{\methodname{}}
        & \textbf{0.9441 $\pm$ 0.0017}
        & \textbf{0.8728 $\pm$ 0.0050}
        & \textbf{0.9309 $\pm$ 0.0024} \\
        \bottomrule
    \end{tabular}
\end{table}


\subsection{Downstream Video Editing}
\label{appen:downstream_editing}
\paragraph{Experimental setup}
\yr{We follow the evaluation metrics of IVEBench~\citep{chen2026ivebench}: Instruction Satisfaction (IS), Phrase Semantic Consistency (PSC), Overall Semantic Consistency (OSC), Subject Consistency (SC), Background Consistency (BC), Temporal Flickering (TF), Motion Smoothness (MS), Video Training Suitability Score (VTSS), Content Fidelity (CF), and Semantic Fidelity (SF). We evaluate all 64 \textit{Subject Motion Editing} instances from the Short subset. All methods receive the same source video and edit instruction. RF-Edit~\citep{wang2025taming} and Wan-Edit~\citep{li2025five} additionally use the benchmark source and target prompts required by their respective editing pipelines. For \methodname{}, given a source video, we first recover a reconstruction-oriented inverse prompt and minimally rewrite it according to the edit instruction using GPT-4o~\citep{hurst2024gpt}. We then apply an
additional GPT-4o verification pass to ensure that the requested motion is reflected in the rewritten prompt, revising it if necessary. The edited prompt and only the first frame of the source video are then used for video generation.}

\paragraph{Subcategory analysis}
\yr{We additionally report results for the two predefined Subject Motion Editing subcategories in IVEBench: single-subject and multiple-subject motion editing. As shown in Table~\ref{appen:editing_by_motion}, \methodname{} achieves the highest Instruction Satisfaction and Content Fidelity in both subcategories, consistent with the aggregate results in Table~\ref{tab:editing_results}.}
\begin{table}[h]
    \centering
    \caption{
        \yr{\textbf{Results by editing subcategory.}
        Results on the two predefined Subject Motion Editing subcategories of IVEBench~\citep{chen2026ivebench}.}
    }
    \label{appen:editing_by_motion}
    \small
    \setlength{\tabcolsep}{5pt}

    \begin{subtable}[t]{\textwidth}
        \centering
        \caption{\textbf{Single-subject motion editing.}}
        \label{appen:editing_single}

        \begin{tabular}{llccc}
            \toprule
            \textbf{Dimension} & \textbf{Metric}
            & \textbf{RF-Edit}
            & \textbf{Wan-Edit}
            & \textbf{MVIP (Ours)} \\
            \midrule

            \multirow{3}{*}{Compliance}
            & Instruction Satisfaction
            & 2.94 & 1.88 & \textbf{3.06} \\
            & Phrase Semantic Consistency
            & \textbf{0.19} & \textbf{0.19} & \textbf{0.19} \\
            & Overall Semantic Consistency
            & 0.24 & \textbf{0.25} & \textbf{0.25} \\

            \midrule
            \multirow{5}{*}{Quality}
            & Subject Consistency
            & 0.92 & \textbf{0.93} & \textbf{0.93} \\
            & Background Consistency
            & 0.95 & \textbf{0.96} & \textbf{0.96} \\
            & Temporal Flickering
            & \textbf{0.98} & 0.97 & \textbf{0.98} \\
            & Motion Smoothness
            & \textbf{0.99} & 0.98 & \textbf{0.99} \\
            & Video Training Suitability Score
            & 0.03 & \textbf{0.05} & \textbf{0.05} \\

            \midrule
            \multirow{2}{*}{Fidelity}
            & Content Fidelity
            & 3.09 & 3.79 & \textbf{3.91} \\
            & Semantic Fidelity
            & 0.79 & \textbf{0.98} & 0.96 \\
            \bottomrule
        \end{tabular}
    \end{subtable}

    \vspace{0.8em}

    \begin{subtable}[t]{\textwidth}
        \centering
        \caption{\textbf{Multiple-subject motion editing.}}
        \label{appen:editing_multiple}

        \begin{tabular}{llccc}
            \toprule
            \textbf{Dimension} & \textbf{Metric}
            & \textbf{RF-Edit}
            & \textbf{Wan-Edit}
            & \textbf{MVIP (Ours)} \\
            \midrule

            \multirow{3}{*}{Compliance}
            & Instruction Satisfaction
            & 3.27 & 2.33 & \textbf{3.47} \\
            & Phrase Semantic Consistency
            & 0.18 & \textbf{0.19} & \textbf{0.19} \\
            & Overall Semantic Consistency
            & 0.23 & \textbf{0.25} & 0.24 \\

            \midrule
            \multirow{5}{*}{Quality}
            & Subject Consistency
            & 0.89 & \textbf{0.92} & \textbf{0.92} \\
            & Background Consistency
            & 0.93 & \textbf{0.95} & \textbf{0.95} \\
            & Temporal Flickering
            & \textbf{0.98} & 0.96 & 0.97 \\
            & Motion Smoothness
            & \textbf{0.99} & 0.98 & \textbf{0.99} \\
            & Video Training Suitability Score
            & 0.02 & \textbf{0.05} & \textbf{0.05} \\

            \midrule
            \multirow{2}{*}{Fidelity}
            & Content Fidelity
            & 3.03 & 3.83 & \textbf{4.00} \\
            & Semantic Fidelity
            & 0.79 & \textbf{0.96} & 0.95 \\
            \bottomrule
        \end{tabular}
    \end{subtable}
\end{table}

\paragraph{Robustness to random seeds}
\yr{
We evaluate the robustness of \methodname{} to diffusion noise by using three independently sampled random seeds per video. As shown in Table~\ref{appen:editing_random_seed}, \methodname{} maintains performance comparable to the default-seed setting with consistently small standard deviations across metrics, indicating that its effectiveness does not depend on using the same
seed for reconstruction and editing. These results further support the practical applicability of \methodname{} without explicit seed control.
}
\begin{table}[h]
    \centering
    \small
    \caption{
        \yr{\textbf{Robustness to random seeds in downstream video editing.}
        Random-seed results are reported as mean $\pm$ standard deviation
        over three independently sampled seeds per video.}
    }
    \label{appen:editing_random_seed}

    \begin{tabular}{@{}llcccc@{}}
        \toprule
        \textbf{Dimension}
        & \textbf{Metric}
        & \shortstack{\textbf{RF-Edit}\\{\scriptsize\citep{wang2025taming}}}
        & \shortstack{\textbf{Wan-Edit}\\{\scriptsize\citep{li2025five}}}
        & \shortstack{\textbf{\methodname{}}\\\textbf{(Default)}}
        & \shortstack{\textbf{\methodname{}}\\\textbf{(Random)}} \\
        \midrule

        Compliance
        & Instruction Satisfaction
        & 3.09
        & 2.09
        & 3.25
        & \textbf{3.302 $\pm$ 0.039} \\

        & Phrase Semantic Consistency
        & 0.18
        & \textbf{0.19}
        & \textbf{0.19}
        & \textbf{0.191 $\pm$ 0.001} \\

        & Overall Semantic Consistency
        & 0.23
        & \textbf{0.25}
        & \textbf{0.25}
        & 0.247 $\pm$ 0.001 \\

        \midrule

        Quality
        & Subject Consistency
        & 0.90
        & \textbf{0.93}
        & 0.92
        & 0.921 $\pm$ 0.002 \\

        & Background Consistency
        & 0.94
        & \textbf{0.95}
        & \textbf{0.95}
        & \textbf{0.951 $\pm$ 0.001} \\

        & Temporal Flickering
        & \textbf{0.98}
        & 0.97
        & 0.97
        & \textbf{0.976 $\pm$ 0.001} \\

        & Motion Smoothness
        & \textbf{0.99}
        & 0.98
        & \textbf{0.99}
        & 0.986 $\pm$ 0.001 \\

        & Video Training Suitability Score
        & 0.02
        & \textbf{0.05}
        & \textbf{0.05}
        & \textbf{0.052 $\pm$ 0.001} \\

        \midrule

        Fidelity
        & Content Fidelity
        & 3.06
        & 3.81
        & \textbf{3.95}
        & 3.932 $\pm$ 0.045 \\

        & Semantic Fidelity
        & 0.79
        & \textbf{0.97}
        & 0.96
        & 0.956 $\pm$ 0.001 \\

        \bottomrule
    \end{tabular}
\end{table}

\paragraph{Additional qualitative results}
\yr{
We further provide qualitative analyses of downstream video editing. As shown in Figure~\ref{appen:additional_editing}, the compared editing methods are applied to the same source video under the same edit instruction. Table~\ref{appen:editing_prompts} further compares the prompts before and after editing for the examples shown in Figures~\ref{fig:editing_results} and Figure~\ref{appen:additional_editing}, respectively.
}
\begin{center}

\includegraphics[
    width=\textwidth,
    trim={0 120 0 90},
    clip
]{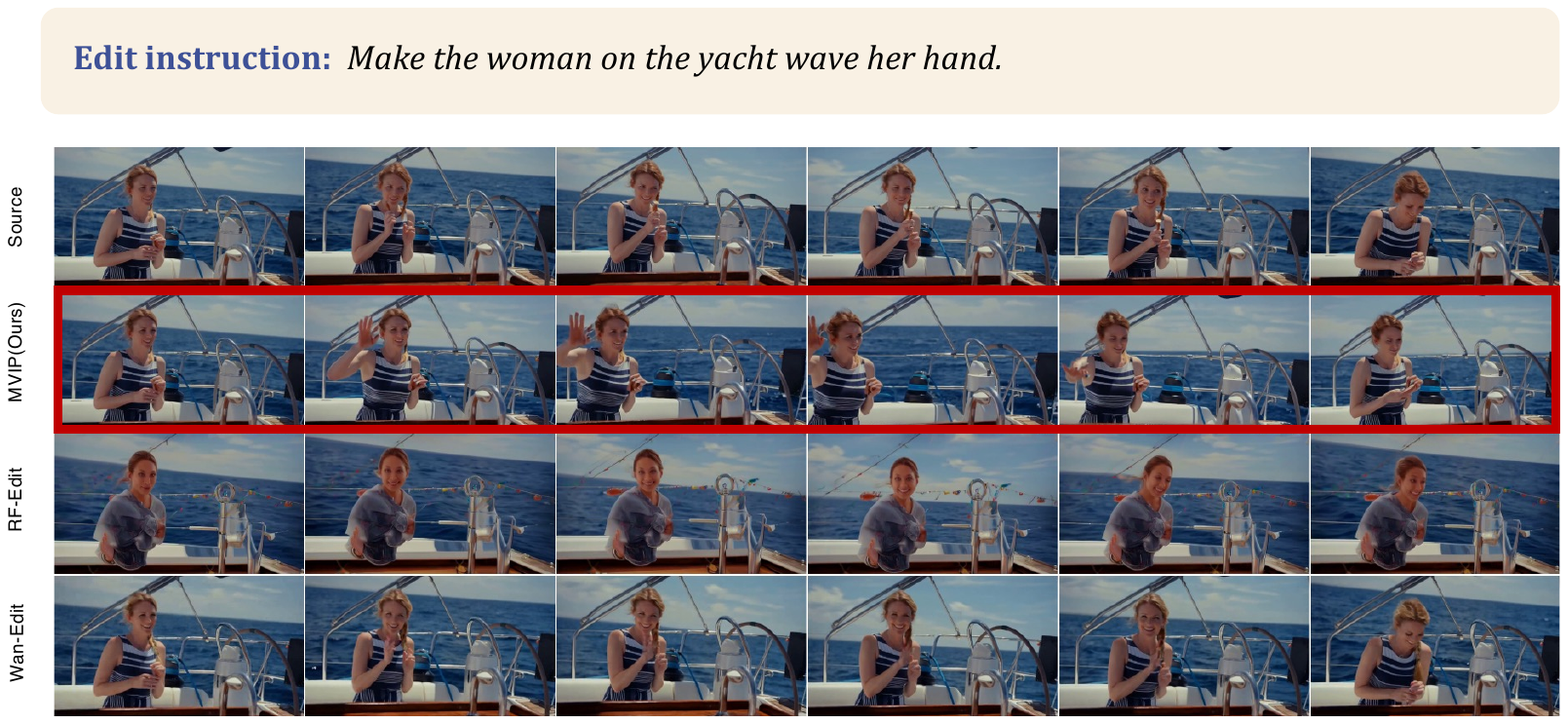}

\textbf{(a)}

\includegraphics[
    width=\textwidth,
    trim={0 120 0 90},
    clip
]{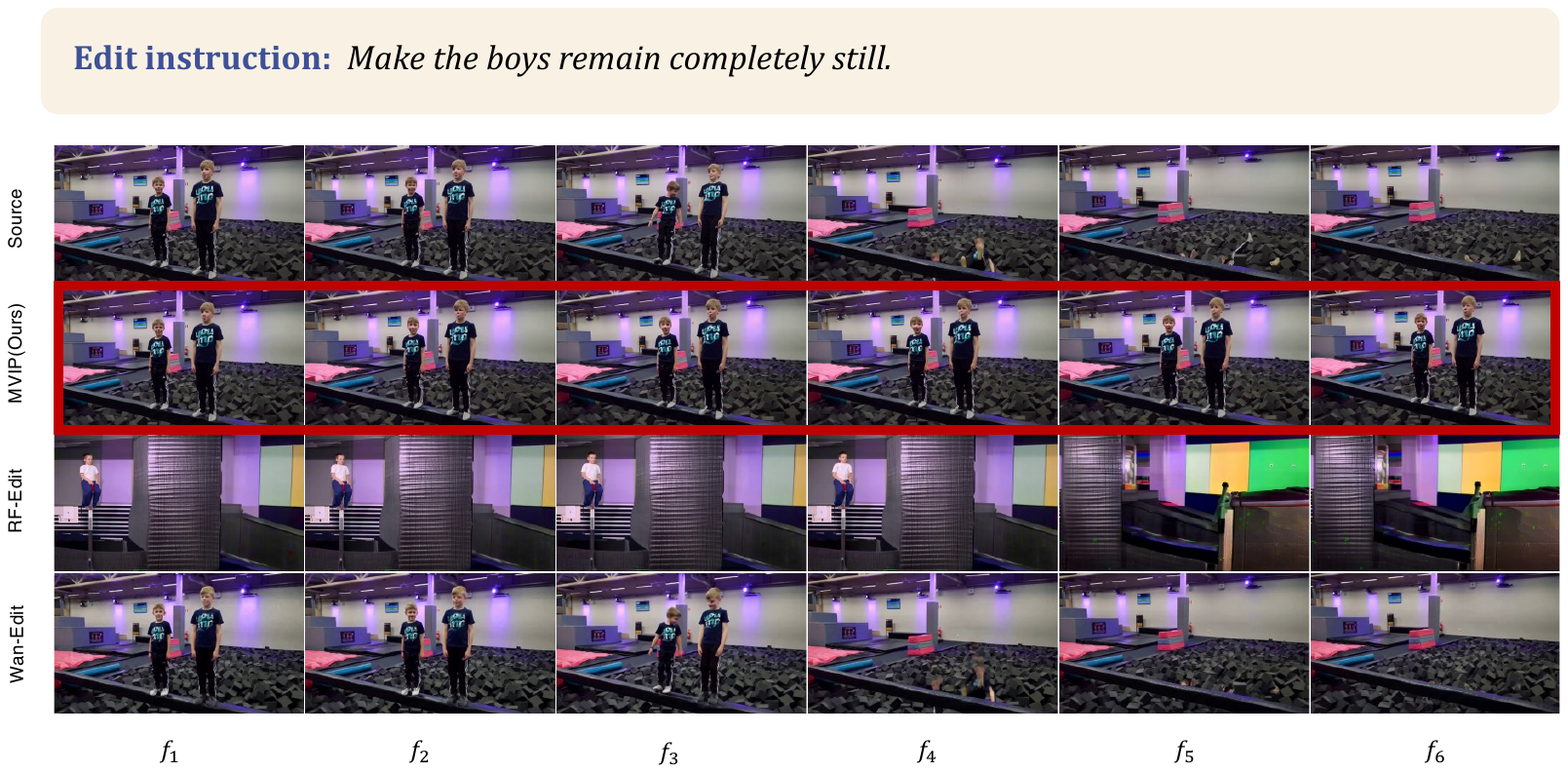}

\textbf{(b)}

\includegraphics[
    width=\textwidth,
    trim={0 120 0 90},
    clip
]{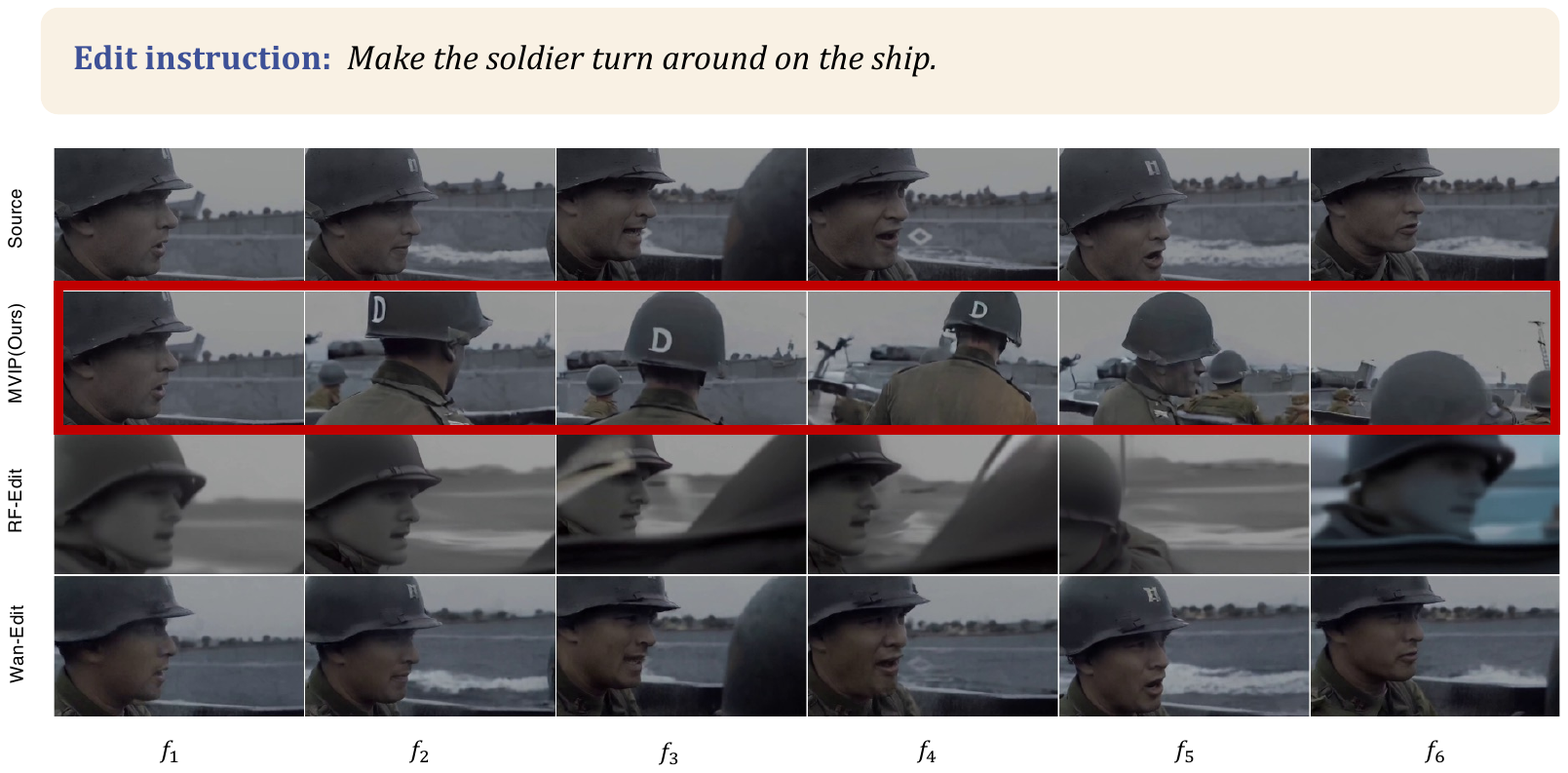}

\textbf{(c)}

\includegraphics[
    width=\textwidth,
    trim={0 120 0 90},
    clip
]{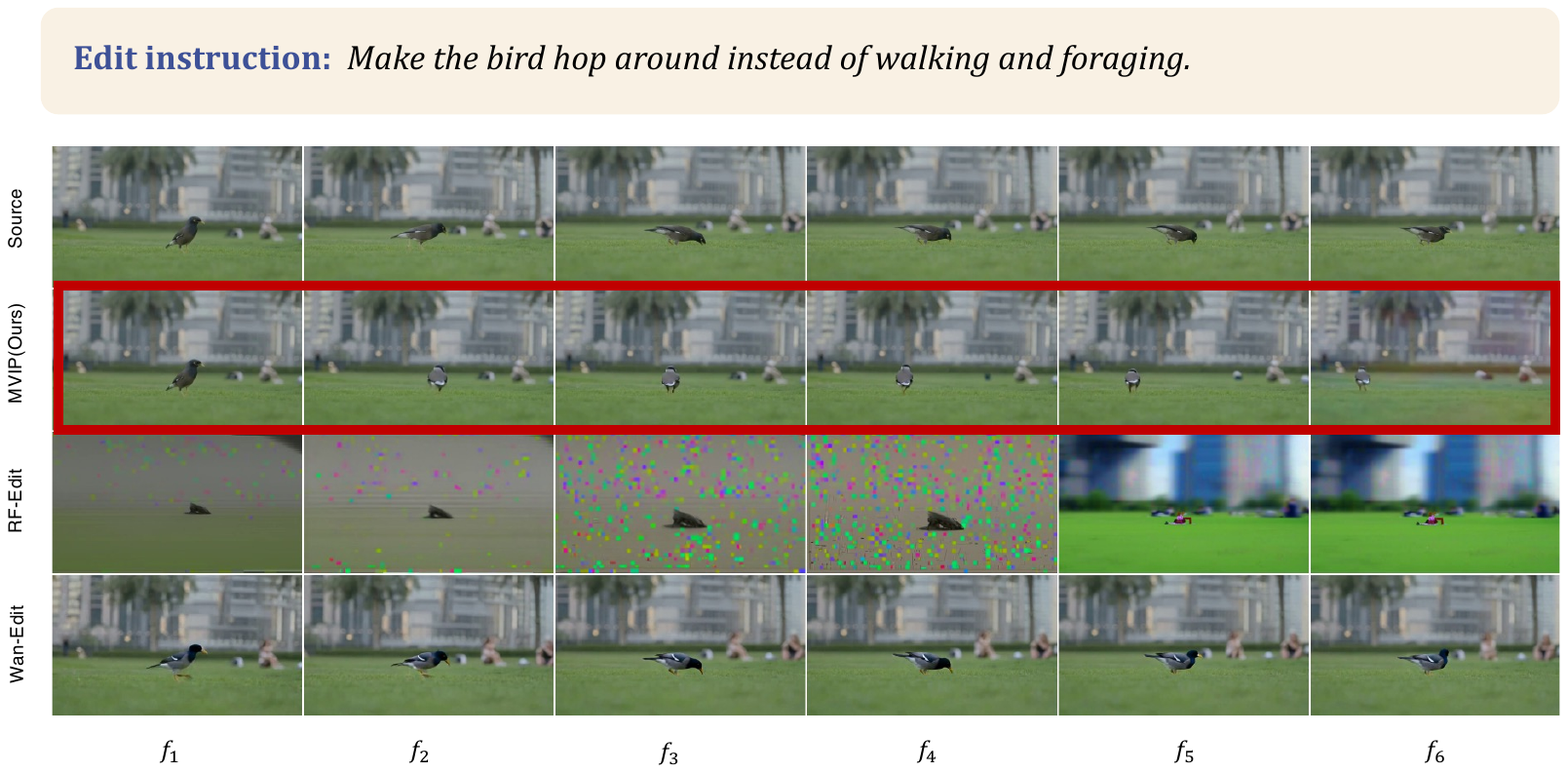}

\textbf{(d)}

\captionof{figure}{
    \yr{\textbf{Qualitative comparison on downstream video editing.}}
}
\label{appen:additional_editing}

\end{center}

\clearpage
\begin{table}[H]
    \centering
    \caption{
        \textbf{Prompt transformations for downstream video editing.}
        Relevant prompt changes are highlighted in bold.
    }
    \label{appen:editing_prompts}

    \footnotesize
    \setlength{\tabcolsep}{5pt}
    \renewcommand{\arraystretch}{1.15}

    \begin{subtable}[t]{\textwidth}
        \centering
        \caption{
            Edit instruction: ``Make the man and the boy suddenly fall asleep.''
        }
        \label{appen:editing_prompts_main}

        \begin{tabularx}{\textwidth}{
            @{}
            p{0.14\textwidth}
            >{\raggedright\arraybackslash}X
            >{\raggedright\arraybackslash}X
            @{}
        }
            \toprule

            \textbf{Method}
            &
            \hfill
            \shortstack{
                \textbf{Before editing}\\
                \scriptsize Source / Recovered
            }
            \hfill\mbox{}
            &
            \hfill
            \shortstack{
                \textbf{After editing}\\
                \scriptsize Target / Edited
            }
            \hfill\mbox{}
            \\

            \midrule

            \textbf{IVEBench}
            &
            A man and a boy are sitting at a table in a room with floral
            wallpaper. \textbf{The man is holding a phone and a bottle of
            beer, while the boy is facing away from the camera, looking at
            an open book on the table.}
            &
            A man and a boy are sitting at a table in a room with floral
            wallpaper. \textbf{The man and the boy suddenly fall asleep.}
            \\

            \addlinespace

            \textbf{MVIP (Ours)}
            &
            The person on the left, presumably male based on attire and body
            structure, is seated casually, wearing a maroon polo shirt with a
            logo or emblem printed on it, paired with beige shorts shorts.
            His posture indicates attentiveness, as he
            \textbf{leans slightly toward the table, suggesting engagement
            in the ongoing discourse}. \ldots{} The right, another individual,
            presumably younger, clad in a plaid, button-down,
            \textbf{partakes actively albeit subtlyly, his back angled toward
            us, indicating introspection or evasion}.
            &
            The person on the left, presumably male based on attire and body
            structure, is seated casually, wearing a maroon polo shirt with a
            logo or emblem printed on it, paired with beige shorts shorts.
            His posture indicates attentiveness, as he leans slightly toward
            the table, \textbf{but he suddenly falls asleep}, suggesting
            engagement in the ongoing discourse. \ldots{} The right, another
            individual, presumably younger, clad in a plaid, button-down,
            partakes actively albeit subtlyly, his back angled toward us,
            \textbf{but he suddenly falls asleep}, indicating introspection
            or evasion.
            \\

            \bottomrule
        \end{tabularx}
    \end{subtable}

    \vspace{0.8em}


\begin{subtable}[t]{\textwidth}
    \centering
    \caption{
        Edit instruction: ``Make the boys remain completely still.''
    }
    \label{appen:editing_prompts_0221}

    \begin{tabularx}{\textwidth}{
        @{}
        p{0.14\textwidth}
        >{\raggedright\arraybackslash}X
        >{\raggedright\arraybackslash}X
        @{}
    }
        \toprule

        \textbf{Method}
        &
        \hfill
        \shortstack{
            \textbf{Before editing}\\
            \scriptsize Source / Recovered
        }
        \hfill\mbox{}
        &
        \hfill
        \shortstack{
            \textbf{After editing}\\
            \scriptsize Target / Edited
        }
        \hfill\mbox{}
        \\

        \midrule

        \textbf{IVEBench}
        &
        Two boys are standing on a balance beam in an indoor trampoline
        park. They \textbf{jump off the beam into a foam pit filled with
        black foam blocks}.
        &
        Two boys are standing on a balance beam in an indoor trampoline
        park, \textbf{remaining completely still in place, without jumping
        into the foam pit filled with black foam blocks}.
        \\[3pt]

        \textbf{MVIP (Ours)}
        &
        The video shows two young boys \textbf{standing} on a foam pit,
        which is a large area filled with foam blocks designed for jumping
        and landing safely. They appear to be in an indoor recreational or
        sports center, evidenced the surrounding equipment and the general
        setup of the space
        &
        The video shows two young boys \textbf{remaining completely still}
        on a foam pit, which is a large area filled with foam blocks
        designed for jumping and landing safely. They appear to be in an
        indoor recreational or sports center, evidenced the surrounding
        equipment and the general setup of the space
        \\

        \bottomrule
    \end{tabularx}
\end{subtable}

\end{table}

\subsection{VBench}
We further evaluate the generated videos using VBench~\cite{huang2024vbench},
which measures reference-free video generation quality along multiple dimensions,
including subject consistency, background consistency, temporal flickering, motion smoothness, and dynamic degree. As shown in Table \ref{appen:vbench}, \methodname{} achieves the best or competitive scores across most dimensions on both real and synthetic videos. In particular, \methodname{} achieves the highest background consistency on both splits and the highest subject consistency on synthetic videos, while remaining competitive on real videos, suggesting that the recovered prompts help maintain coherent subject appearance and scene context during generation. While Captioning+Refine slightly improves flicker and motion smoothness in some cases, \methodname{} maintains strong temporal quality while producing
more semantically consistent videos. These results suggest that the proposed inverse prompts improve reconstruction-oriented alignment without sacrificing
general video quality.

\begin{table}[h]
\centering
\caption{\textbf{VBench results on real and synthetic videos.}}
\label{appen:vbench}
\resizebox{\linewidth}{!}{
\begin{tabular}{lccccc}
\toprule
\textbf{Method} & \textbf{Subject} $\uparrow$ & \textbf{Background} $\uparrow$ & \textbf{Flicker} $\uparrow$ & \textbf{Motion Smooth.} $\uparrow$ & \textbf{Dynamic} $\uparrow$ \\
\midrule
\multicolumn{6}{l}{\textbf{Real videos}} \\
PEZ~\cite{wen2023hard} & \textbf{0.7832} & 0.8897 & 0.9774 & 0.9855 & 0.8448 \\
Captioning (Qwen2.5-VL)~\cite{bai2025qwen25vl} & 0.7706 & 0.8892 & 0.9788 & 0.9871 & 0.8793 \\
Captioning + Refine~\cite{ren2025reverse} & 0.7716 & 0.8907 & 0.9786 & 0.9880 & \textbf{0.8966} \\
\textbf{MVIP (Ours)} & 0.7777 & \textbf{0.8919} & \textbf{0.9816} & \textbf{0.9888} & 0.8621 \\
\midrule
\multicolumn{6}{l}{\textbf{Synthetic videos}} \\
PEZ~\cite{wen2023hard} & 0.9286 & 0.9571 & 0.9792 & 0.9885 & 0.4375 \\
Captioning (Qwen2.5-VL)~\cite{bai2025qwen25vl} & 0.9153 & 0.9502 & 0.9738 & 0.9829 & \textbf{0.4875} \\
Captioning + Refine~\cite{ren2025reverse} & 0.9289 & 0.9576 & \textbf{0.9836} & \textbf{0.9901} & 0.3750 \\
\textbf{MVIP (Ours) }& \textbf{0.9335} & \textbf{0.9581} & 0.9826 & \textbf{0.9901} & 0.4500 \\
\bottomrule
\end{tabular}
}
\end{table}

\subsection{Image-text alignment}
\label{appen:it_align}
\begin{table}[t]
\centering
\caption{
\textbf{Text-alignment evaluation.}
}
\label{app:text_alignment_table}

\small
\setlength{\tabcolsep}{6pt}
\renewcommand{\arraystretch}{1.08}

\begin{tabular}{llccc}
\toprule
Split & Method
& ViCLIP V-T$\uparrow$
& SigLIP$\uparrow$
& BLIP2-ITM$\uparrow$ \\
\midrule
\multirow{5}{*}{Synthetic}
& FLF-Only & N/A & N/A & N/A \\
& PEZ~\cite{wen2023hard} & 0.1820 & 0.0711 & 0.5050 \\
& Captioning (Qwen2.5-VL)~\cite{bai2025qwen25vl} & 0.2687 & 0.1408 & 0.5213 \\
& Captioning + Refine~\cite{ren2025reverse} & 0.2644 & 0.1419 & 0.5172 \\
& \textbf{MVIP (Ours)} & \textbf{0.2814} & \textbf{0.1550} & \textbf{0.7986} \\
\midrule
\multirow{5}{*}{Real}
& FLF-Only & N/A & N/A & N/A \\
& PEZ~\cite{wen2023hard} & 0.1718 & 0.0649 & 0.3551 \\
& Captioning (Qwen2.5-VL)~\cite{bai2025qwen25vl} & 0.2675 & 0.1231 & 0.4426 \\
& Captioning + Refine~\cite{ren2025reverse} & 0.2663 & 0.1203 & 0.4182 \\
& \textbf{MVIP (Ours)} & \textbf{0.2708} & \textbf{0.1485} & \textbf{0.7124} \\
\bottomrule
\end{tabular}
\end{table}

Table \ref{app:text_alignment_table} reports additional video-text and image-text alignment metrics. This evaluation complements the video-level reconstruction metrics by measuring whether the recovered prompt is semantically aligned with the target video. The FLF-only baseline is marked as N/A because it relies only on first-last-frame conditioning and does not provide a text prompt to evaluate. Across both synthetic and real videos, \methodname{} achieves the highest scores on all
metrics. These results indicate that the proposed method recovers prompts that better capture the semantic content of the target video, rather than merely improving visual or temporal similarity.

\subsection{Analysis of the Learned Soft Prompt}
\label{appen:soft_prompt_analysis}
\paragraph{\yr{Effect on generated prompts}}
We further examine the effect of the learned soft prompt $\Phi$ in
Figure~\ref{fig:soft_prompt_ablation}. When sampled without the learned $\Phi$, \textit{i.e.}, the target VLM is
conditioned only on the manually written hard meta-prompt, the generated prompt mainly captures the coarse action while omitting details about the subject, scene context, and temporal progression. With $\Phi$, on the other hand, the prompt significantly better specifies the subject's appearance, the surrounding environment, and how the motion unfolds over time. This comparison shows that $\Phi$ acts as a learned prompt prior, steering the VLM toward visual and temporal cues that are useful for the video inversion process. 

\begin{figure}[h]
    \centering

    \includegraphics[width=0.95\linewidth]{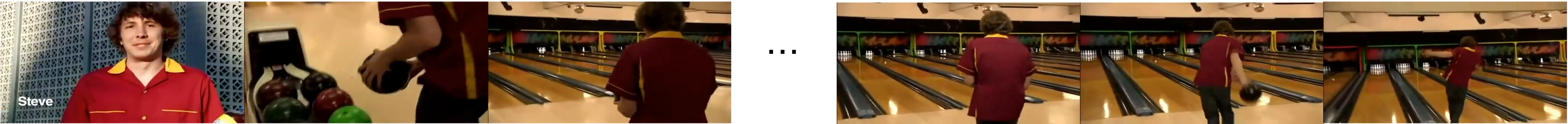}

    \vspace{0.5em}

    \scriptsize
    \setlength{\tabcolsep}{5pt}
    \renewcommand{\arraystretch}{1.18}

    \begin{tabularx}{0.95\linewidth}{@{}p{0.16\linewidth}X@{}}
        \toprule
        \textbf{Setting} & \textbf{Generated Prompt} \\
        \midrule

        Without $\Phi$ &
        C throws the bowling ball. \\
        
        \midrule

        With $\Phi$ (\textbf{Ours}) &
        A young man in a maroon and yellow bowling shirt is standing on the bowling lane,
        preparing to bowl. He is holding a black bowling ball in his right hand and is in the process of winding up for his throw. The bowling alley is well-lit with overhead
        lights, and there are other people in the background, some of whom are also bowling.
        The lanes are clearly marked, and the pins are set up at the end of the lane. The young man appears focused and determined as he prepares to release the ball down
        the lane. \\
        \bottomrule
    \end{tabularx}

    \caption{
    \textbf{Qualitative comparison of prompts generated without and with the learned soft prompt $\Phi$.}
    The learned soft prompt yields a more detailed reconstruction-oriented prompt, capturing the
    subject, scene context, and action progression, while the prompt without $\Phi$ only describes
    the coarse action.
    }
    \label{fig:soft_prompt_ablation}
    \vspace{-0.15in}
\end{figure}

\paragraph{\yr{Natural-language interpretation}}
\yr{To better understand what the learned soft prompt encodes, we perform a
qualitative analysis following the soft-prompt-to-instruction conversion used
in InstructZero~\citep{chen2024instructzero}. For each checkpoint, we provide
the soft prompt together with the same set of video--caption input-output examples
and ask an LLM to infer the underlying instruction. As shown in Table~\ref{tab:soft_prompt_decode}, the decoded outputs evolve from a generic
clarification request to a task-specific description of video captioning,
suggesting that the soft prompt gradually acquires task-level semantics during
meta-training. These outputs should be viewed as approximate interpretations
rather than exact discrete inversions of the soft prompt.}

\begin{table}[h]
    \centering
    \small
    \caption{
    \yr{
    \textbf{Natural-language interpretation of the learned soft prompt
    during meta-training.}
    We decode the continuous soft prompt at different training steps
    using the procedure of InstructZero~\citep{chen2024instructzero}.
    }}
    \label{tab:soft_prompt_decode}
    \begin{tabular}{c p{0.78\linewidth}}
        \toprule
        \textbf{Training step} & \textbf{{Decoded output (InstructZero)~\citep{chen2024instructzero}}} \\
        \midrule
        30 &
        ``It seems like you've provided a series of unrelated video clips
        and descriptions. Could you please clarify what specific information
        you're looking for?'' \\
        \addlinespace
        60 &
        ``The prompt you provided seems to be a series of unrelated video
        clips with different scenes and themes. It appears to be asking for
        a description or analysis of the content shown in each clip.'' \\
        \addlinespace
        100 &
        ``It seems like you've provided a series of video clips with
        different scenes and themes. It looks like the task was to describe
        what each video clip shows.'' \\
        \bottomrule
    \end{tabular}
\end{table}

\subsection{Efficiency}
Table~\ref{tab:reptile_inner_timing} reports the computational cost of increasing the inner-loop adaptation depth. Using three inner-loop steps increases the wall-clock time per outer update from 10.65 minutes to 33.01 minutes, corresponding to a 3.10$\times$ higher runtime. Since the additional inner-loop steps provide limited performance gain in Table~\ref{tab:phi_initialization}, we use $K=1$ as our default setting for better efficiency.

\begin{table}[h]
\centering
\caption{
\textbf{Runtime analysis of inner-loop adaptation depth.}
We report the average wall-clock time per outer update over three matched outer updates under the same training video schedule.
Increasing the number of inner-loop adaptation steps from $K=1$ to $K=3$ increases runtime by 3.10$\times$.
}
\label{tab:reptile_inner_timing}

\small
\setlength{\tabcolsep}{6pt}
\renewcommand{\arraystretch}{1.08}

\begin{tabular}{ccc}
\toprule
Inner steps $K$ & Time / outer update $\downarrow$ & Relative cost \\
\midrule
1 & 10.65 min & 1.00$\times$ \\
3 & 33.01 min & 3.10$\times$ \\
\bottomrule
\end{tabular}
\end{table}

\end{document}